%% file: main.tex
\documentclass[10pt,twocolumn,letterpaper]{article}

\usepackage{iccv}              

\input{preamble}

\definecolor{iccvblue}{rgb}{0.21,0.49,0.74}
\usepackage[pagebackref,breaklinks,colorlinks,allcolors=iccvblue]{hyperref}

\def\paperID{10399} 
\def\confName{ICCV}
\def\confYear{2025}

\title{\textit{Revelio}: Interpreting and leveraging semantic information in diffusion models}

\newcommand{\sigmalabel}{$\sigma_{label}$}
\newcommand{\sdone}{SD 1.5}
\newcommand{\sdtwo}{SD 2.1}
\newcommand{\oxford}{Oxford-IIIT Pet}
\newcommand{\caltech}{Caltech-101}
\newcommand{\aircraft}{FGVC-Aircraft}
\newcommand{\imagenet}{ImageNet}
\newcommand{\modelName}{Diff-C}

\author{
Dahye Kim$^{1}$\thanks{Equal contribution.}\quad\quad  Xavier Thomas$^{1}$\footnotemark[1]\quad\quad Deepti Ghadiyaram$^{12}$\thanks{Corresponding author.} \\
$^1$Boston University \quad
$^2$Runway \\
{\tt\small {\{dahye, xthomas, dghadiya\}@bu.edu} }}

\begin{document}
\maketitle
\input{sec/0_abstract}    
\input{sec/1_intro}
\input{sec/2_relatedwork}
\input{sec/3_approach}
\input{sec/4_experiments}
\input{sec/5_conclusion}

\clearpage
\noindent\textbf{Acknowledgments.} We thank Jonathan Granskog and our fellow research group members at BU for helpful discussions and paper feedback.

{
    \small
    \bibliographystyle{ieeenat_fullname}
    \bibliography{main}
}

\input{sec/X_suppl}

\end{document}

%% file: preamble.tex
\usepackage{listings}
\usepackage[accsupp]{axessibility}
\usepackage{multirow}  
\usepackage{booktabs}  
\usepackage{graphicx}  
\usepackage{tcolorbox}
\usepackage{listings}  
\usepackage{graphicx}  
\newcommand{\increase}[1]{#1~\textcolor{green}{$\uparrow$}}
\newcommand{\decrease}[1]{#1~\textcolor{red}{$\downarrow$}}
\usepackage[normalem]{ulem}

\usepackage{caption}

%% file: sec/0_abstract.tex
\begin{abstract}
     We study \textit{how} rich visual semantic information is represented within various layers and denoising timesteps of different diffusion architectures. We uncover monosemantic interpretable features by leveraging k-sparse autoencoders (k-SAE). We substantiate our mechanistic interpretations via transfer learning using light-weight classifiers on off-the-shelf diffusion models' features. On $4$ datasets, we demonstrate the effectiveness of diffusion features for representation learning. We provide an in-depth analysis of how different diffusion architectures, pre-training datasets, and language model conditioning impacts visual representation granularity, inductive biases, and transfer learning capabilities. Our work is a critical step towards deepening interpretability of black-box diffusion models. Code and visualizations available at: \url{https://github.com/revelio-diffusion/revelio}
\end{abstract}

%% file: sec/1_intro.tex
\section{Introduction}
\label{sec:intro}
Generating high-quality photo-realistic and creative visual content using diffusion models is a thriving area of research. For a generative model to accurately simulate the visual world around us, its latent space should in principle capture rich visual semantics and the physical dynamics of the real world. A direct empirical evidence is in recent efforts that leverage diffusion features for discriminative tasks such as detection~\cite{diffusion_detect}, segmentation~\cite{label_seg, diffusion_segmentation}, classification~\cite{diffusion_classifier_23}, semantic correspondence~\cite{hyperfeatures}, depth estimation~\cite{wu2023datasetdm, zhao2023unleashing}, or visual reasoning~\cite{diva} tasks. Yet, they do not offer clear insights on \textit{how} this rich semantic information is represented within the model. Some prior attempts that visualize attention maps~\cite{ban2024understanding} or use PCA~\cite{plugandplay} on the intermediate features, though valuable, operate on per-image basis and thus, do not offer a more holistic in-depth interpretation of diffusion models' internal representations.

In this work, we go beyond harnessing the rich representations of diffusion models and aim to fundamentally understand and interpret diffusion models' internal states. Concretely, we address the following questions: what flavors of visual information is captured in different layers and time-steps of a diffusion model? How do they interact with and complement each other and the overall learnt visual information? Do different layers benefit differently from external conditioning and why? What inductive biases are uniquely captured in convolution-based diffusion models compared to transformer-based ones? 
\input{fig_latex/main}

Understanding how a model learns visual information offers several key benefits. First, current visual generative models are black box in nature: it is not clear why a benign prompt sometimes produces an unsafe output or why a very slight tweak to the same prompt generates a very different output~\cite{openai_jailbreak}. Answering the above fundamental questions will be a crucial step towards interpreting black box generative models. Second, distilling the granularity of semantic information represented across different layers, timesteps, and model architectures can aid in designing more efficient algorithms that offer semantic and style control. 

To \textit{reveal} the visual knowledge learnt by diffusion models, we adopt ``mechanistic interpretation'' techniques and learn a sparse dictionary of monosemantic visual concepts. It is physiologically proven that human visual system \textit{sparsely} encodes the most recurring visual patterns using a small set of basis functions~\cite{sparse_encoding}. Motivated by this, we aim to uncover interpretable features by leveraging k-sparse auto-encoders (k-SAE)~\cite{ksparse}, which have been shown to help interpret language models~\cite{sparse_lang, sparse_anthropic}. We illustrate \textit{how} the semantic visual information is packed differently depending on the representation granularity of the test dataset, across different diffusion layers, denoising timesteps, model architectures, and pre-training data.
\input{fig_latex/fig2}
Going beyond this, we corroborate our mechanistic interpretation by learning very light-weight classifiers on top of off-the-shelf diffusion models' features. Through rigorous analysis against multiple baselines and benchmarks, we show the surprising effectiveness of diffusion features across a variety of tasks: coarse and fine-grained classification and complex visual reasoning. Unlike all prior works, our classifier, dubbed, \modelName, bypasses the need to employ additional losses~\cite{diffusion_classifier_23}, training a student model~\cite{diffusion_distillation}, or training a feature map fusing method~\cite{hyperfeatures}, thereby offering significant computational benefits (\textbf{4} orders of magnitude inference speedup compared to \cite{diffusion_classifier_23}\footnote{Our method requires training a classifier to achieve this.}). We summarize our empirical and interpretable analysis below, which align perfectly across datasets, tasks, and model architectures.
\begin{itemize}
 \item \textbf{Representation granularity varies non-linearly with model depth}, with different diffusion layers capturing varying levels of visual semantic information, from coarse-grained shape, texture, or local color patterns to fine-grained animal breed details, to more global visual concepts like camera angles and object poses.
 \item \textbf{Representation granularity and generalizability varies with diffusion architectures, pre-training data, latent or pixel space, cross and self-attention mechanisms} -- design choices made to improve the overall pixel generation quality and training efficiency.
 \item \textbf{k-sparse autoencoders help isolate monosemantic visual properties} systematically across model states and help interpret black box diffusion models.
\end{itemize}

%% file: fig_latex/main.tex
\begin{figure}[t]
    \centering
    {\includegraphics[width=1\linewidth]{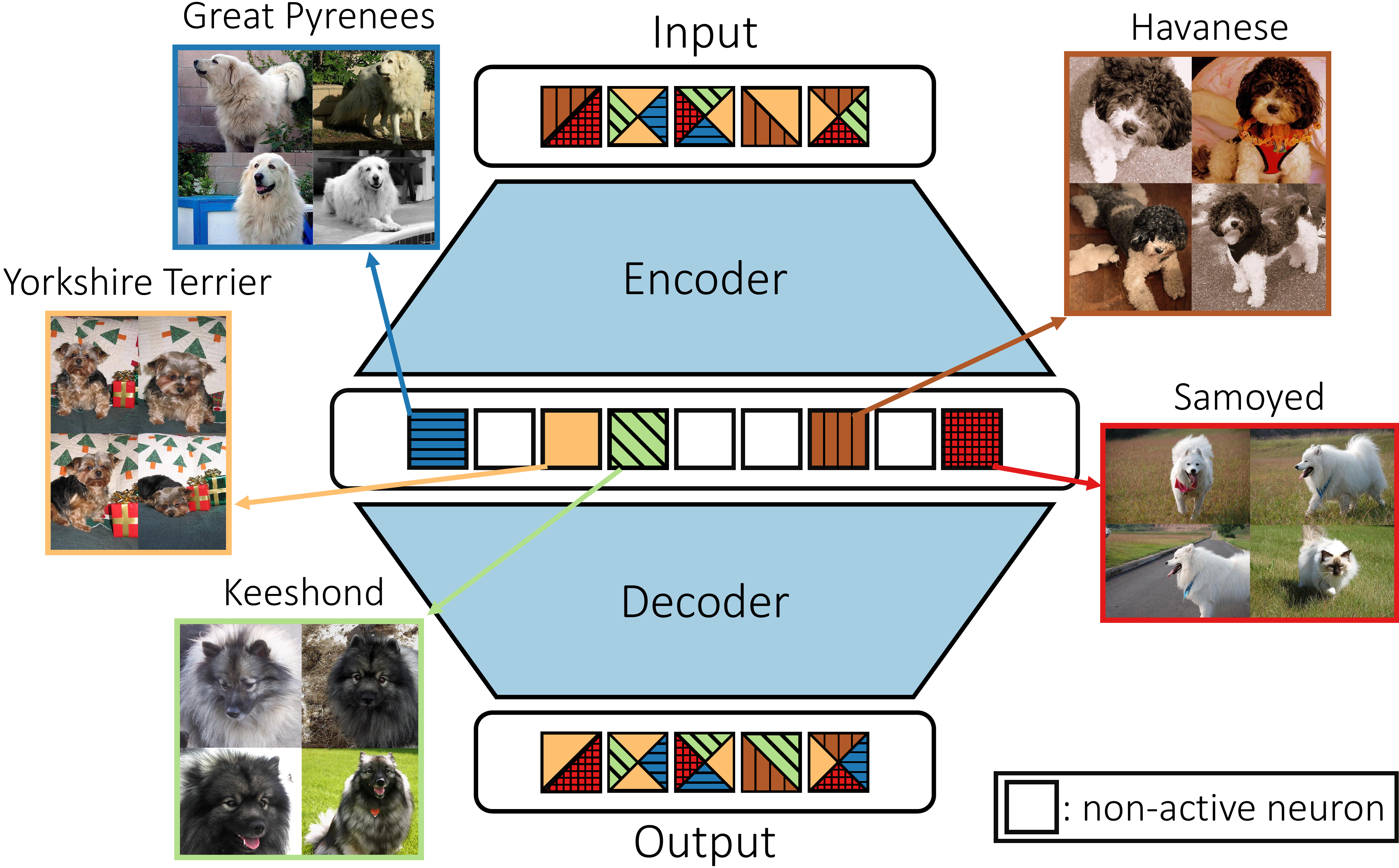}}\hfill \\ 
    \caption{\footnotesize{\textbf{k-sparse autoencoders (k-SAE)} trained on complex visual features help identify monosemantic visual properties represented within black-box diffusion models. We show sample k-SAE neurons and top-4 images that yield highest activations when the k-SAE is trained on intermediate diffusion layer's features on \oxford~\cite{oxford} dataset. Note how these features encapsulate distinct fine-grained information about different breeds like \textit{Keeshond} and \textit{Samoyed}. Best viewed in color.
    } }\label{fig:main}
    \vspace{-0.2in}
\end{figure} 

%% file: fig_latex/fig2.tex
\begin{figure*}[t]
    \centering
{\includegraphics[width=0.7\textwidth]{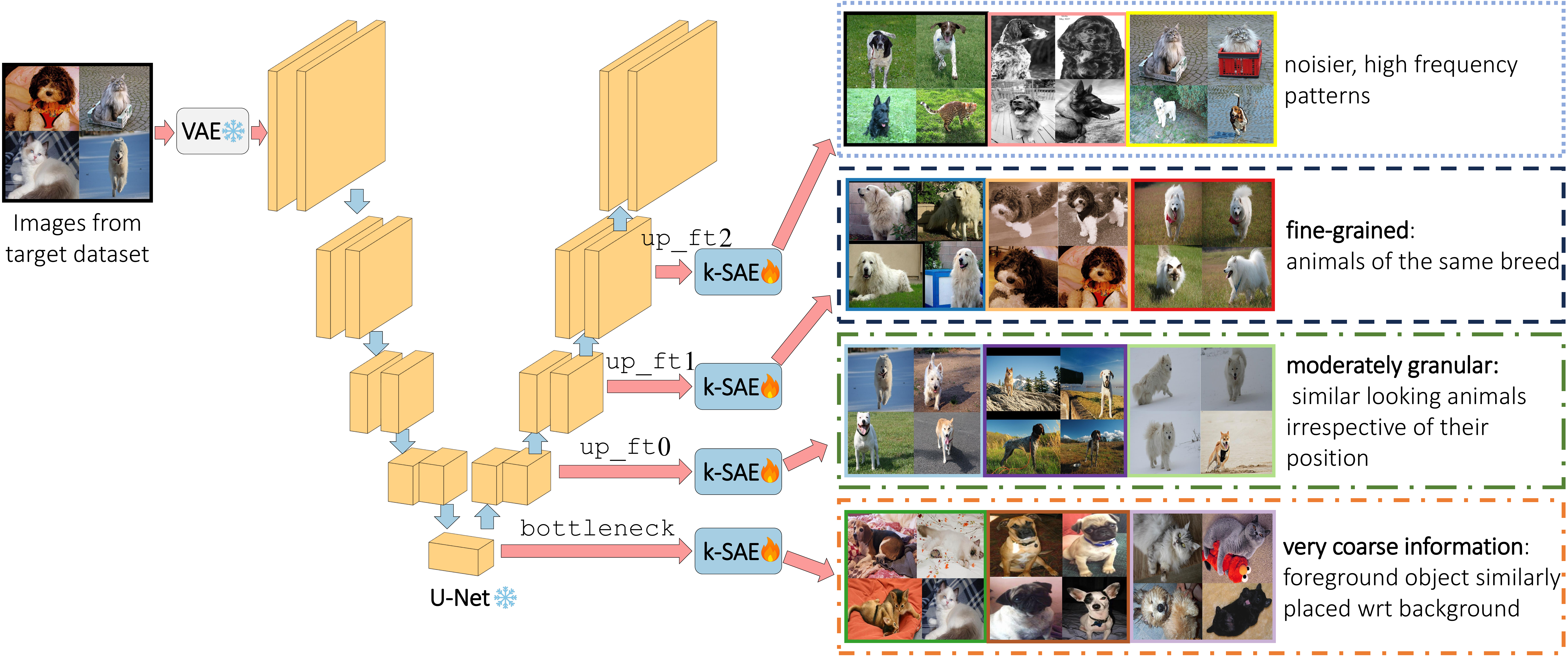}}
    \caption{\footnotesize{\textbf{k-SAE visualizations across layers of the U-Net} in {\sdone} and sample images from different neurons yielding highest activations when k-SAEs are trained on different layers for $t=25$ on \oxford. We note that across $3$ random neurons of k-SAEs, the \texttt{bottleneck} layer captures very coarse-grained information, where foreground objects positioned similarly are activated by the same neuron. \texttt{up\_ft1} captures valuable breed specific domain information while \texttt{up\_ft2} seems to capture high-frequency visual patterns.}}\label{fig:diffusion_unet}
    \vspace{-0.1in}
\end{figure*}

%% file: sec/2_relatedwork.tex
\section{Related Work}
\noindent \textbf{{Diffusion features for discriminative tasks:}} Diffusion models have achieved remarkable results in generating semantically rich high-resolution images~\cite{ho2020denoising, dhariwal2021diffusion, stablediffusion, imagen, bao2023all, dit}. Several recent works leverage diffusion features beyond image and video synthesis: for zero-shot classification~\cite{clark2024text, diffusion_classifier_23, diffusion_distillation}, detection~\cite{diffusion_detect}, segmentation~\cite{diffusion_segmentation, label_seg}, semantic correspondence~\cite{hyperfeatures, discffusion}, rendering novel views~\cite{featurenerf}, image editing and semantic image manipulation tasks~\cite{plugandplay, asyrp}, and so on. Our work is different from prior works in two important ways. First, we propose a simple method to \textit{adapt} diffusion features for discriminative tasks without the need to distill~\cite{diffusion_distillation}, train an expensive hyper-network~\cite{hyperfeatures}, or generate synthetic data~\cite{jahanian2021generative}. Second, we go beyond leveraging diffusion features and interpret \textit{how} visual information is packed with the model's architecture.

\noindent \textbf{Interpreting diffusion features:} Some recent studies aim at understanding and interpreting diffusion models~\cite{dewan2024diffusion, basu2024mechanistic, chefer2023hidden}. Plug-and-Play~\cite{plugandplay} performs PCA analysis on intermediate features of Stable Diffusion~\cite{stablediffusion} and finds that intermediate features reveal localized semantic information shared across objects, while early layers capture high-frequency details. However, their analysis is based only on $20$ real and generated humanoid images, limiting the generalizability of their findings to different domains and model architectures. Authors of ~\cite{guo2024dit} explore how diffusion features vary with the underlying architecture. Similarly, the effect of cross-attention layers to image attributes~\cite{voynov2023p+} and semantic information at different timesteps~\cite{mahajan2024prompting, patashnik2023localizing} have been studied by progressively conditioning text prompts. While valuable, these analyses are also done on a per-image basis and do not offer a holistic and in-depth interpretation of the models' internal states. Diffusion Lens~\cite{toker2024diffusion} analyzes the text encoder of diffusion models by generating images from its intermediate representations. By contrast, our work mechanistically interprets the opaque \textit{visual} diffusion features when conditioned on blank prompts using sparse autoencoders. 

Recent works have demonstrated that sparse autoencoders (SAE) could recover monosemantic features in large language models (LLMs) ~\cite{bricken2023monosemanticity, sparse_lang, gao2024scaling} and CLIP vision features~\cite{fry2024towards, daujotas2024interpreting}. Concurrent work~\cite{surkov2024unpacking} investigates the possibility of using SAEs to learn interpretable features from residual updates within the U-Net to investigate how the cross-attention layer integrates the input text prompt. By contrast, our focus is to understand how visual information is packed within the diffusion models' internal states and the interplay between representation abstraction and model design choices. We propose a method to reveal valuable human interpretable visual patterns baked within black box diffusion models.

%% file: sec/3_approach.tex
\section{Approach}
Our goal is to interpret and expand our understanding of black box diffusion models. We address this from two different perspectives: first, we train k-sparse autoencoders~\cite{ksparse} to recover interpretable monosemantic visual semantic features across different layers, timesteps, and diffusion architectures. Second, we substantiate each interpretability finding by training light-weight classifiers on the exact same diffusion features. 

We begin by providing an overview of diffusion models (Sec.~\ref{sec3.1}), followed by motivation and architecture used for training k-SAE (Sec.~\ref{sec3.2}), followed by the light-weight classifier, \textbf{\modelName} (Sec.~\ref{sec3.3}). 
\subsection{Preliminaries on diffusion models}\label{sec3.1}
Several powerful open and close-sourced diffusion models have emerged just in the last two years~\cite{ramesh2022hierarchical, imagen,stablediffusion, sdxl, gen3a, mj, emu_video, dit}. Broadly, diffusion models are probabilistic generative models that aim to learn a data distribution $p(x)$ through an iterative denoising process. During the forward diffusion process, the input image $x$ is gradually perturbed with noise over $T$ timesteps. The reverse process consists of iterative denoising steps, where each step estimates the added noise $\epsilon_{\theta}(x_t, t)$, parameterized by $\theta$, with $t = 1, \dots, T$.
Each iteration takes a noisy image $x_t$ as input and predicts the added noise $\epsilon$. The objective of the diffusion model is given by:
\begin{equation}
L_{DM} = \mathbb{E}_{x,t,\varepsilon\sim\mathcal{N}(0,1)} \left[ \|\varepsilon - \varepsilon_\theta(x_t, t)\|_2^2 \right]
\end{equation}
Instead of operating on images $x$, latent diffusion models (LDM)~\cite{stablediffusion} operate on a latent representation $z$, obtained by mapping the image into a lower-dimensional space using a variational autoencoder~\cite{kingma2013auto} which consists of an encoder $\mathcal{E}$ and decoder $\mathcal{D}$. The diffusion process models the distribution of these latent embeddings, allowing for more efficient computation. The revised objective is:
\begin{equation}
L_{LDM} = \mathbb{E}_{\mathcal{E}(x),t,\varepsilon\sim\mathcal{N}(0,1)} \left[ \|\varepsilon - \varepsilon_\theta(z_t, t)\|_2^2 \right]
\end{equation}
\subsection{Preliminaries on k-sparse autoencoders}\label{sec3.2}
Our aim is to gain insight into how visual information is encapsulated in a diffusion model. Given the very high non-linearity and complex architectures of generative models, identifying interpretable components directly from layer activations is not viable. In this work, we isolate monosemantic features by training k-sparse autoencoders (k-SAEs) on the activations from different diffusion layers, timesteps, and architectures, which we describe next.

Sparse autoencoders~\cite{ng2011sparse} are neural networks to learn compact feature representations in an unsupervised manner. They contain an encoder and a decoder, and are trained with a sparsity penalty and a sample reconstruction loss to encourage only a few neurons to be maximally activated for a given input. However, the sparsity penalty term in SAEs presents significant training challenges~\cite{tibshirani1996regression,ksparse}. A k-sparse autoencoder is an extension of sparse autoencoder~\cite{ng2011sparse} designed to improve the training challenges by explicitly regulating the number of active neurons during training to $k$. Specifically, in each training step, a top-k activation function is used to retain only the $k$ largest neuron activations, while zeroing out the rest. 

Let $x$ denote the $d$-dimensional spatially-pooled diffusion activations\footnote{For notational simplicity, we describe our setup for an arbitrary layer and denoising timestep, but the same method applies for activations from any model state.}. Let $W_{enc} \in \mathbb{R}^{n \times d}$ and $W_{dec} \in \mathbb{R}^{d \times n}$ denote the weight matrices of the k-SAE's encoder and decoder respectively (Fig.~\ref{fig:main}), where $n$ denotes the dimension of the autoencoder's hidden layer. $n$ is equal to $d$ multiplied by a positive integer, called the expansion factor. Following~\cite{bricken2023monosemanticity}, $b_{pre} \in \mathbb{R}^{d}$ denotes the bias term added to input $x$ before feeding to the encoder (aka pre-encoder bias), while $b_{enc} \in \mathbb{R}^{n}$ denotes the bias term for encoder. Upon passing $x$ through the encoder, we obtain $z$ defined as:
\begin{equation}
    z =\text{TopK}(W_{enc}(x-b_{pre})+b_{enc}), \\
\end{equation}
where the $\text{TopK}$ activation function retains only the top $k$ neuron activations and sets the rest to zero~\cite{ksparse}. The decoder then reconstructs $z$, given by:
\begin{equation}
    \hat{x} = W_{dec} z + b_{pre} \\ 
\end{equation}
The training loss is the normalized reconstruction mean squared error (MSE) between the reconstructed feature ($\hat{x}$) and the original feature ($x$), given by:
\begin{equation}
    L_{mse}= \Vert x - \hat{x} \Vert ^2_2
\end{equation}
As we show in results (Sec.~\ref{sec:experiments}), k-SAE plays a key role in qualitatively interpreting visual semantic information. 
\subsection{Diffusion Classifier (\modelName)}~\label{sec3.3}
Next, to quantitatively study the visual semantic information packed within pre-trained diffusion models, we design a lightweight classifier called \textbf{\modelName} to adapt diffusion features to downstream tasks. \textbf{\modelName} (shown in Table~\ref{tab:diffc_unet} in suppl. material) comprises a series of convolutional layers to progressively reduce the spatial dimensions of the diffusion features, followed by a pooling and a downstream task-dependent fully-connected layer. Despite the inherently unique architectures of convolutions-based U-Net~\cite{ronneberger2015u} and diffusion-based DiT~\cite{dit}, we adapt the outputs of different U-Net layers and DiT blocks into 2D feature maps and process them using \modelName.

%% file: sec/4_experiments.tex
\input{sec/expts_setup}

\input{sec/expts_layers}
\input{sec/expts_timesteps}

\input{sec/expts_architectures}
\input{sec/expts_visualreasoning}

\input{sec/expts_sota}

%% file: sec/expts_setup.tex
\section{Experiments} \label{sec:experiments}
In this section, we first share the implementation details and training setup, followed by detailed analyses while dissecting diffusion models. 
\subsection{Implementation details} \label{sec:sub_implement}
Unless otherwise stated, we use Stable Diffusion (SD) 1.5 model~\cite{stablediffusion}, DDIM scheduler~\cite{DDIM}, and an empty prompt used as the text conditioning.

\noindent \textbf{k-SAE:}
For training a k-SAE, we empirically found that $k=32$ yields the best results for different datasets, based on training stability and overall sparsity. Diffusion activations are extracted by passing images from a target dataset into the VAE of a pretrained diffusion model. For SD, we spatially pool diffusion activations resulting in $d = 1280$ for \texttt{bottleneck}, \texttt{up\_ft0}, and \texttt{up\_ft1} layers. 

\noindent \textbf{Diff-C} has $4$ convolutional layers ($conv1$-$conv4$) as shown in Table~\ref{tab:diffc_unet}. The final feature dimension is $1024$. For all classification tasks and datasets, we train on a NVIDIA RTX A6000 GPU, use a batch size of $16$, optimize using AdamW~\cite{adamw}, a learning rate of $1 \times 10^{-4}$. We train \textbf{Diff-C} for $30$ epochs with cosine annealing learning rate schedule and set a minimum learning rate $\eta_{\text{min}}$ ($5 \times 10^{-5}$). We randomly crop and resize input images to $512 \times 512$ and augment with random horizontal flip transformations.
\subsection{Setup} \label{sec:setup}
\noindent \textbf{Datasets and tasks:} We interpret and analyze diffusion features of on four image datasets and against competitive baselines on two tasks: (a) \textbf{classification} Transfer learning onto \oxford~\cite{oxford}, \aircraft~\cite{Maji2013FineGrainedVC}, and \caltech~\cite{caltech}, (b) \textbf{visual reasoning} as in~\cite{eyeshut}, we interleave two visual features with CLIP: DINO~\cite{Oquab2023DINOv2LR} and diffusion features. We use the LLaVA-Lightning~\footnote{We use LLaVA-Lightning due to compute constraints.} configuration and MPT-7B-Chat ~\cite{MosaicML2023Introducing} as the base language model. We use CC595k~\cite{sharma-etal-2018-conceptual} for stage 1 pre-training and LLaVA-Instruct-80K~\cite{llava} for stage 2 fine-tuning (more details in suppl. material).

\noindent \textbf{Notation}: We focus on interpreting the bottleneck and the decoder layers of the UNet, as the information from the encoder is fed into the decoder through the skip connections. As illustrated in Fig.~\ref{fig:diffusion_unet},
features extracted from a given upsampling (decoder) \texttt{block\_index} are denoted as \texttt{up\_ft\{block\_index\}}. \texttt{bottleneck} refers to the central block which has the smallest spatial resolution in U-Net. Following the standard convention of reverse denoising of diffusion models~\cite{ho2020denoising, DDIM}, $t = 0$ corresponds to the timestep where a final, fully denoised image is achieved and higher values of $t$ represent noisier images, with $t = 1000$ denoting pure noise. For referring to the features from DiT we use \texttt{block\_index} to denote the transformer block from which we extract the features.

\noindent \textbf{Evaluation metrics:} For \textbf{Diff-C}, we report top-1 accuracy for all classification tasks. For the visual reasoning task, we evaluate on the LLaVA-Bench (in-the-wild)~\cite{llava} and MM-Vet benchmarks~\cite{yu2023mm}, where model outputs are scored relative to reference answers generated by text-only GPT-4~\cite{openai2024gpt4technicalreport}. For \textbf{k-SAE}, to quantify the granularity of semantic information captured in diffusion features, we measure how ``pure'' the activated k-SAE neurons are. We do this by measuring the average standard deviation in the class labels (\sigmalabel) of the top-10 most highly activating images among the top 1000 most highly activating features of the learned k-SAE. We stress that the class labels are not used for training but only to measure activated neurons' purity of the k-SAE. We also visualize images which result in highest activation for a given k-SAE neuron. Furthermore, to further reduce subjectivity, we frame the task of quantifying the granularity of semantic information in diffusion features as a multiple-choice question-answering problem for GPT-4o~\cite{gpt4o}. We assess the level of semantic detail captured by different diffusion features (prompt details in suppl. material). We note that GPT-4o predictions can be noisy, hence we primarily relied on label purity (\sigmalabel) for accurate quantification.

%% file: sec/expts_layers.tex
\subsection{Information granularity across diffusion layers}\label{sec:layers}
\input{fig_latex/qual_step25_oxfordpet_up1_up2}
\input{fig_latex/test_acc_comp_layers}
\input{tables_latex/chatgpt}
In this section, we study how the visual semantic information arranges itself across different layers of a pre-trained diffusion network. Specifically, how does the diffusion training objective of minimizing global reconstruction loss impact the visual information granularity across layers? To this end, we extract diffusion features from \texttt{bottleneck}, \texttt{up\_ft0}, and \texttt{up\_ft1}, train separate k-SAE and \textbf{Diff-C} models, and report evaluation metrics listed in Sec.~\ref{sec:setup}.\\
\textbf{Fine-grained classification task:} From Table~\ref{tab:layer}, we note that \texttt{up\_ft1} yields the lowest {\sigmalabel}, indicating that the features corresponding to this layer contain most class-specific information compared to other layers. This is qualitatively corroborated by Fig.~\ref{fig:qual_step25_oxfordpet_up1_up2} where images that k-SAE neurons get most activated by, have very clear class-specific characteristics when using \texttt{up\_ft1} features compared to \texttt{bottleneck} and \texttt{up\_ft2}. Moreover, the accuracy of GPT-4o predictions (Table~\ref{tab:chatgpt}) suggests that \texttt{up\_ft1} captures more fine-grained information, whereas other layers tend to capture moderately granular or very coarse features. This finding is also consistent with \textbf{Diff-C} results presented in Fig.~\ref{fig:cap_comp_layers} for \oxford \ and for another fine-grained dataset: \aircraft~\cite{Maji2013FineGrainedVC}. 
Note that there is a sharp decline in performance at \texttt{up\_ft2} layer and beyond, suggesting that \texttt{up\_ft2} features may be more aligned with the pre-training task objective of pixel reconstruction for image generation, thus are less generalizable for transfer learning. A similar observation was made about the later layers when mechanistically interpreting language models~\cite{logitlens}.
\input{tables_latex/ksae}

\input{fig_latex/qual_step25_caltech_mid_up1}

\noindent \textbf{Does the trend hold for coarse-grained classification task?} To deconflate the effect of task-granularity from diffusion feature granularity, we study the diffusion features from \caltech\ dataset. From Table~\ref{tab:layer}, it is evident that \texttt{bottleneck} features yields a significantly smaller {\sigmalabel}. Also note that the difference between {\sigmalabel} values between layers is quite larger for \caltech\ compared to \oxford.\ To understand this better, we visualize the highest activated images from different layers. From Fig.~\ref{fig:qual_step25_caltech_mid_up1}, we note that those from \texttt{bottleneck} are more class-centric and thus ``purer'' compared to \texttt{up\_ft1}, which may be capturing more style or texture specific information. We hypothesize that for the task of classifying \caltech, coarser shape information is sufficient which is compactly provided by \texttt{bottleneck}. This can be clearly seen from Fig.~\ref{fig:cap_comp_layers}, where, compared to \oxford, the performance gap between \texttt{up\_ft1} and \texttt{bottleneck} is significantly low. However, for challenging tasks where finer-grained information is required, higher-level layers (\texttt{up\_ft1}) are more beneficial.

\noindent \textbf{Does the trend hold for larger datasets?} We conduct the same analysis on ImageNet~\cite{deng2009imagenet}, a relatively coarser-grained dataset. From Table~\ref{tab:layer}, we observe that even on a larger dataset, the \texttt{bottleneck} layer captures more class-specific information, exhibiting similar behavior as on \caltech.

%% file: fig_latex/qual_step25_oxfordpet_up1_up2.tex
\begin{figure}[t]
    \centering
    {\includegraphics[width=0.8\linewidth]{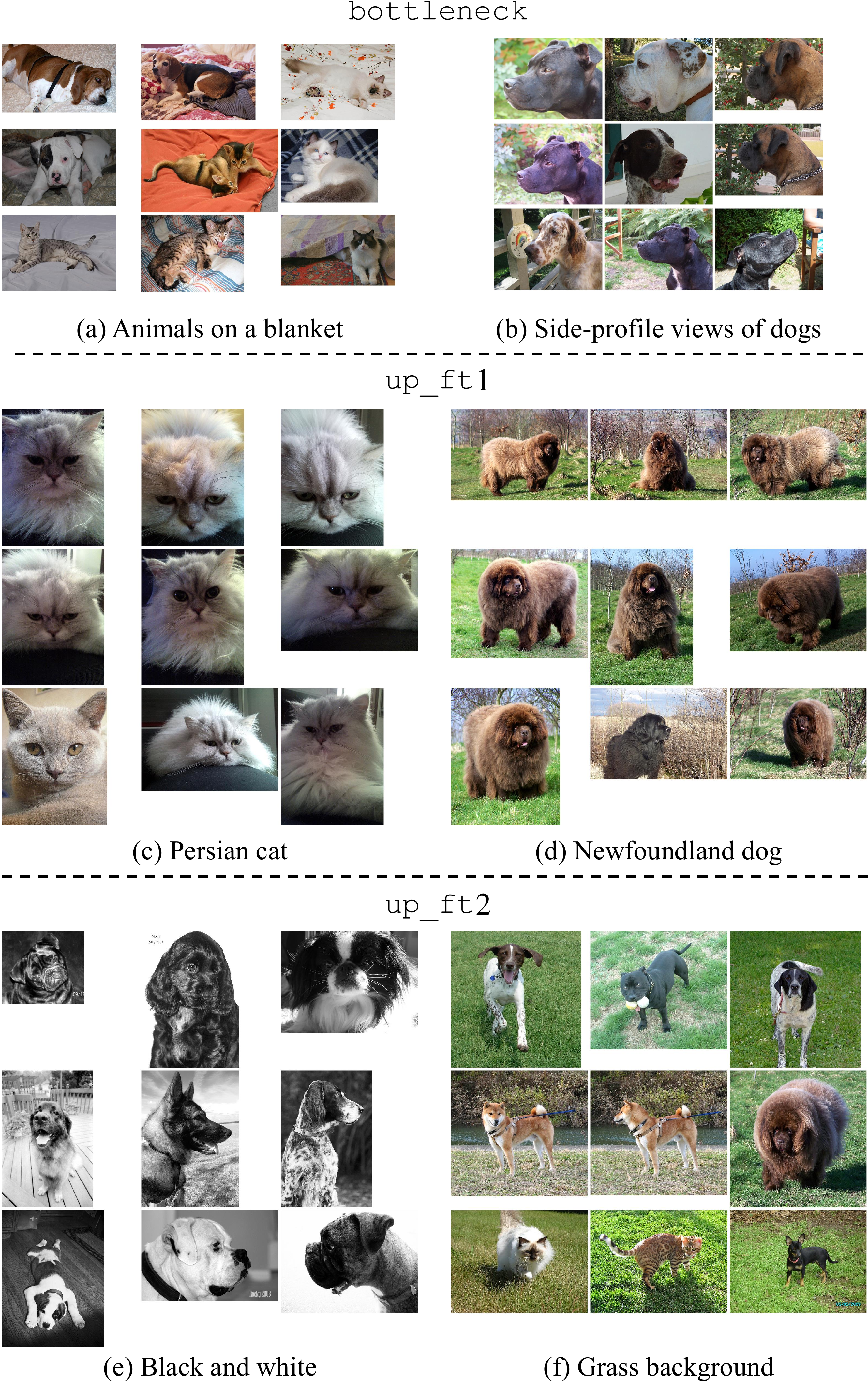}}
    \hspace{-0.02in}
    \caption{\footnotesize{\textbf{k-SAE visualizations on \oxford} of \texttt{bottleneck}, \texttt{up\_ft1}, and \texttt{up\_ft2} U-Net layers at $t=25$.
    \texttt{bottleneck} isolates very coarse patterns of objects positioned similarly with respect to the background. For \texttt{up\_ft1}, clear class-specific features are observed helping us isolate different fine-grained breeds.
    \texttt{up\_ft2} captures more global texture information such as that of grass.}} 
    \label{fig:qual_step25_oxfordpet_up1_up2} 
   \vspace{-0.13in}
\end{figure} 

%% file: fig_latex/test_acc_comp_layers.tex
\begin{figure}[t]
    \vspace{-0in}
    \centering
    {\includegraphics[width=0.7\linewidth]{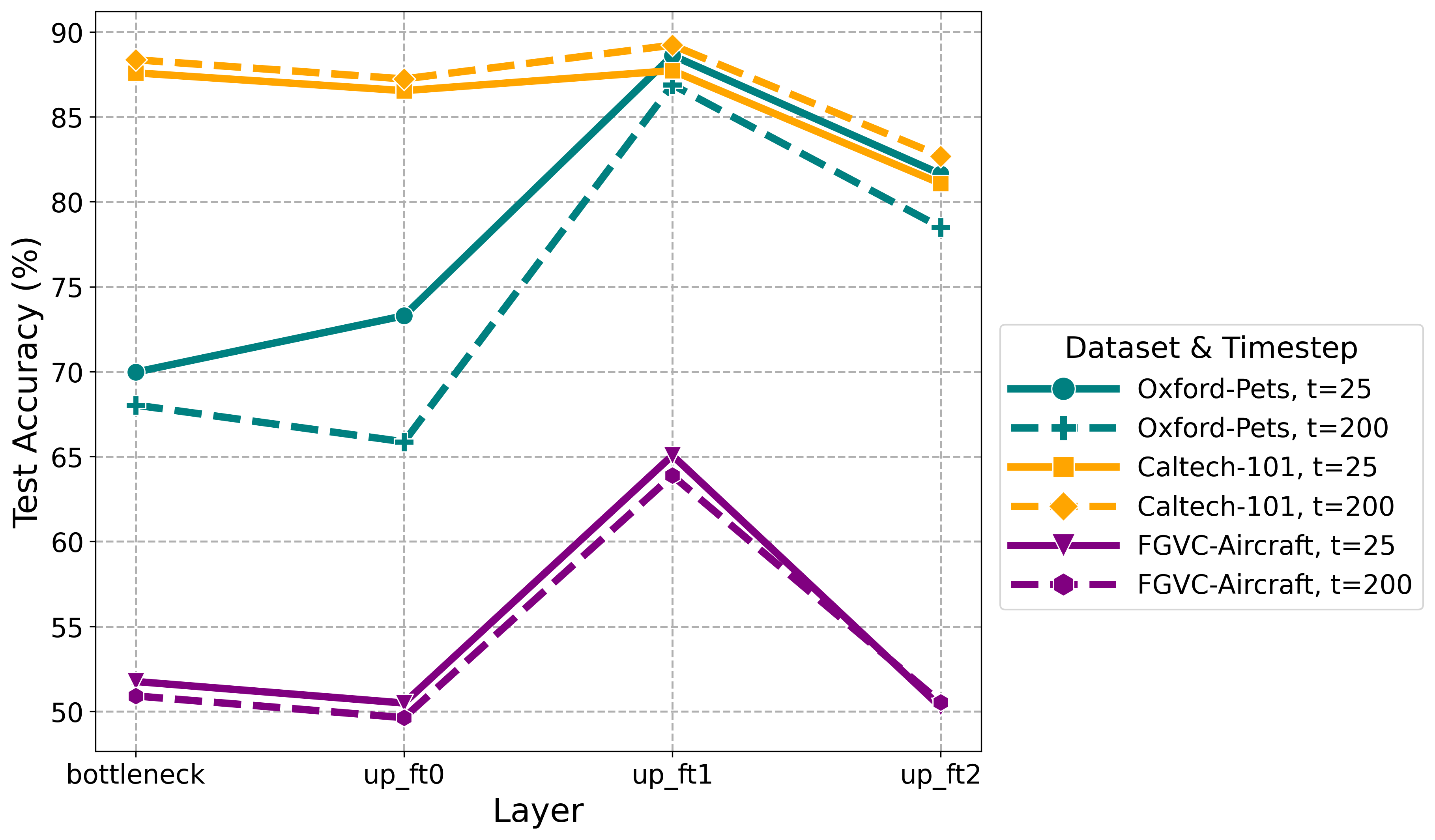}\label{fig:comp_layers}}\hfill \\ 
    \vspace{-0.05in}
    \caption{\footnotesize {\textbf{Top-1 accuracy of different {\sdone} layer features}. Features from {\texttt{up\_ft1}} consistently yield best performance  for {\sdone}.}}\label{fig:cap_comp_layers}
    \vspace{-0.1in}
\end{figure} 

%% file: tables_latex/chatgpt.tex
\begin{table}[t]
    \centering
    \scriptsize
    \renewcommand{\arraystretch}{1.2}
    \resizebox{0.99\linewidth}{!}{
        \centering
        \begin{tabular}{lcccc}
            \textbf{Layer} & \textbf{1. fine-grained}& \textbf{2. Moderately granular} & \textbf{3. Very coarse}  & \textbf{4. No patterns} \\ \hline
            \texttt{bottleneck} & 25 & 52 &  22  & 1\\
            \texttt{up\_ft0}    &33& 46 &  20 & 1\\
            \texttt{up\_ft1}    & \textbf{47} &  41 & 11 & 1\\
            \texttt{up\_ft2}    & 28 & 50 & 20 & 2 \\
        \end{tabular}
        }
   \caption{\footnotesize{\textbf{Accuracy (\%)} of GPT-4o predictions among four multiple-choice options, given the top 10 most highly activating images among the top 100 most highly activating neurons of the k-SAE trained on \oxford. Of the 4 layers, note how \texttt{up\_ft1} features are most fine-grained. We note that GPT-4o predictions can be noisy and very sensitive to system prompt.}}\label{tab:chatgpt}
    \vspace{-0.2in}
\end{table}

%% file: tables_latex/ksae.tex
\begin{table*}[!htbp]
    \centering
    \scriptsize
    \setlength{\tabcolsep}{4pt}
    \renewcommand{\arraystretch}{1.2}
    \resizebox{1\textwidth}{!}{
    \begin{subtable}[t]{0.4\textwidth}
        \centering
        \begin{tabular}{lccc}
            \textbf{Layer} & \textbf{\oxford }& \textbf{\caltech} & \textbf{\imagenet} \\ \hline
            \texttt{bottleneck} & 9.48 & \textbf{9.35} &  \textbf{25.35}  \\
            \texttt{up\_ft0}    & 9.90 & 15.65 &  29.10\\
            \texttt{up\_ft1}    & \textbf{8.59} &  21.33 & 34.3\\
            \texttt{up\_ft2}    & 9.67  & 25.61 & 36.38 \\
            \\
            \\
            \\
        \end{tabular}
        \captionsetup{width=1\textwidth}
        \caption{\textbf{{\sigmalabel} for different layers:} For \oxford, \texttt{up\_ft1} achieves the lowest {\sigmalabel}, whereas \texttt{bottleneck} yields lowest {\sigmalabel} for \caltech, indicating the interplay of representation and task granularity.}
        \label{tab:layer}
    \end{subtable}
    \begin{subtable}[t]{0.38\textwidth}
        \centering
        \begin{tabular}{lcc}
            $t$ & \textbf{\oxford }& \textbf{\caltech}  \\
            & (\textbf{\texttt{up\_ft1}}) & (\textbf{\texttt{bottleneck}}) \\
            \hline
            0   & 8.99          & 11.91 \\
            25  & \textbf{8.59}   & 9.35\\
            100 &8.87           & 8.72 \\
            200 & 8.94          &  \textbf{8.17} \\
            300 & 9.01          & 10.41 \\
            500 &  9.53          &  16.65\\
        \end{tabular}
        \captionsetup{width=0.85\textwidth}
        \caption{\textbf{{\sigmalabel} for different diffusion timesteps:} For \oxford\ \texttt{up\_ft1}, $t = 25$ yields the lowest {\sigmalabel}, whereas for \caltech\ \texttt{bottleneck}, $t = 200$ yields the lowest {\sigmalabel}.}
        \label{tab:timestep}
    \end{subtable}
    \begin{subtable}[t]{0.2\textwidth}
        \centering
        \begin{tabular}{lc}
            \textbf{Model} & \textbf{\oxford } \\\hline
            {\sdone} & \textbf{8.59} \\
            {\sdtwo} & 9.67 \\
            \\
            \\
            \\
            \\
            \\
        \end{tabular}
        \captionsetup{width=0.93\textwidth}
        \caption{\textbf{{\sigmalabel} for {\sdone} vs. {\sdtwo}:} {\sdone} captures more class-specific information than {\sdtwo}.}
        \label{tab:model}
    \end{subtable}
    ~
    \begin{subtable}[t]{0.23\textwidth}
        \centering
        \begin{tabular}{lc}
            \textbf{Block} & \textbf{\oxford} \\\hline
            6 & 10.18 \\
            10 & 9.44 \\
            14 & \textbf{9.05}\\
            18 & 9.55 \\
            22 &9.84 \\
            \\
            \\
        \end{tabular}
        \captionsetup{width=0.85\textwidth}
        \caption{\footnotesize {\textbf{\sigmalabel\ for different DiT blocks} at $t=25$. Mid-blocks of DiT yield the lowest {\sigmalabel} than other layers.}}
        \label{tab:block}
    \end{subtable}
    } 
    \caption{\footnotesize{\textbf{Label purity (\sigmalabel)} measured by computing the average standard deviation {in the class labels (\sigmalabel)} of the top-10 most highly activating images among the top 1000 most highly activating features of the learned k-SAEs for different diffusion layers, timesteps, models, and architectures on \oxford, \caltech and \imagenet.}}\label{tab:ksae}
    \vspace{-0.1in}
\end{table*}

%% file: fig_latex/qual_step25_caltech_mid_up1.tex
\begin{figure}[t]
    \centering
    {\includegraphics[width=1\linewidth]{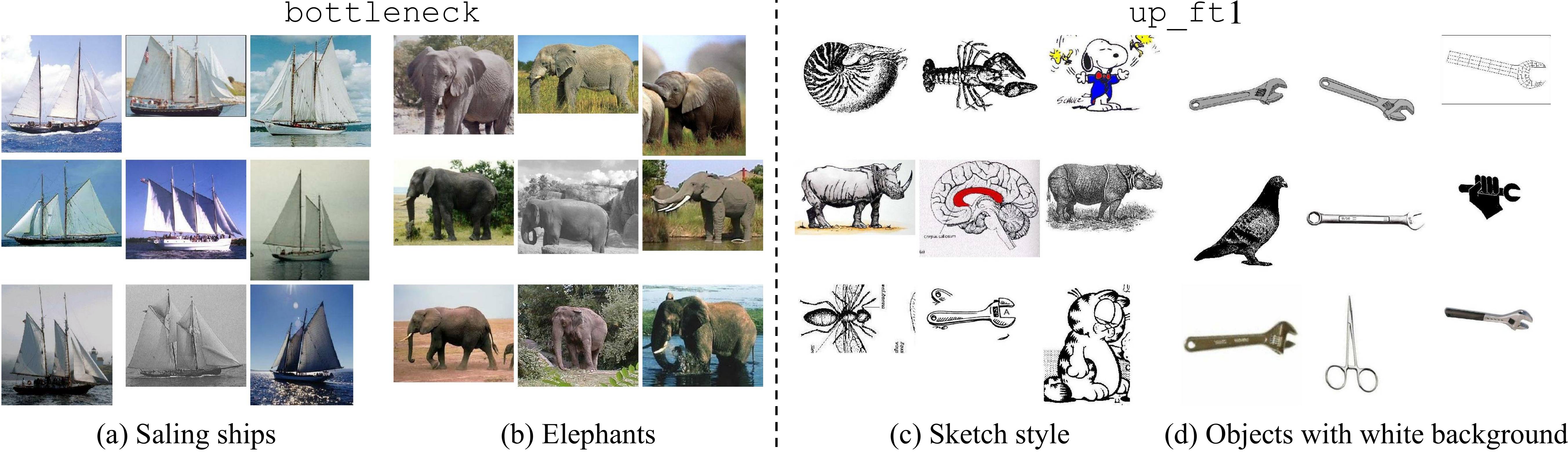}}\hfill \\ 
    \caption{\footnotesize \textbf{k-SAE visualizations on \caltech} of \texttt{bottleneck} and \texttt{up\_ft1} UNet layers at $t=25$.
    Unlike for fine-grained dataset (Fig.~\ref{fig:qual_step25_oxfordpet_up1_up2}), \texttt{bottleneck} captures class information, likely due to distinct object shapes (sailing ships v/s elephants). \texttt{up\_ft1} captures more abstract information such as sketches or objects with white background.} 
   \vspace{-0.2in}
    \label{fig:qual_step25_caltech_mid_up1} 
\end{figure} 

%% file: sec/expts_timesteps.tex
\vspace{-4pt}
\subsection{Information packed across diffusion timesteps}\label{sec:timesteps}
\input{fig_latex/test_acc_trend_pets}

We now examine the interplay between diffusion denoising timesteps and visual semantic information granularity. To this end, we extract diffusion features from \texttt{up\_ft1} at different timesteps $t=\{25, 100, 200, 300, 400, 500\}$, train separate k-SAE and \textbf{Diff-C} models. From Table~\ref{tab:timestep}, we observe that $t=25$ yields the lowest {\sigmalabel}.
This is validated both by top activated images shown in Fig.~\ref{fig:qual_step25_oxfordpet_up1_up2} and \textbf{Diff-C} performance in Fig.~\ref{fig:cap_test_acc_decay}. k-SAE neurons are being activated by images with very clear class-specific characteristics when using features extracted at $t=25$ (more visualizations in suppl. material).
This finding is also consistent with \textbf{Diff-C} results presented in Fig.~\ref{fig:cap_test_acc_decay} for \oxford\ and \aircraft. By contrast, from Table~\ref{tab:timestep}, we find that for \caltech, features extracted at $t = 200$ yield lowest {\sigmalabel}. This finding is also consistent with \textbf{\modelName} results from Fig.~\ref{fig:cap_comp_layers}. This finding is corroborated in~\cite{diffusion_classifier_23, label_seg}. We hypothesize that the additional noise added at $t = 200$ could be helping in making features more generalizable, but deeper investigation is needed in the future.

%% file: fig_latex/test_acc_trend_pets.tex
\begin{figure}[t]
    \centering
    {\includegraphics[width=0.7\linewidth]{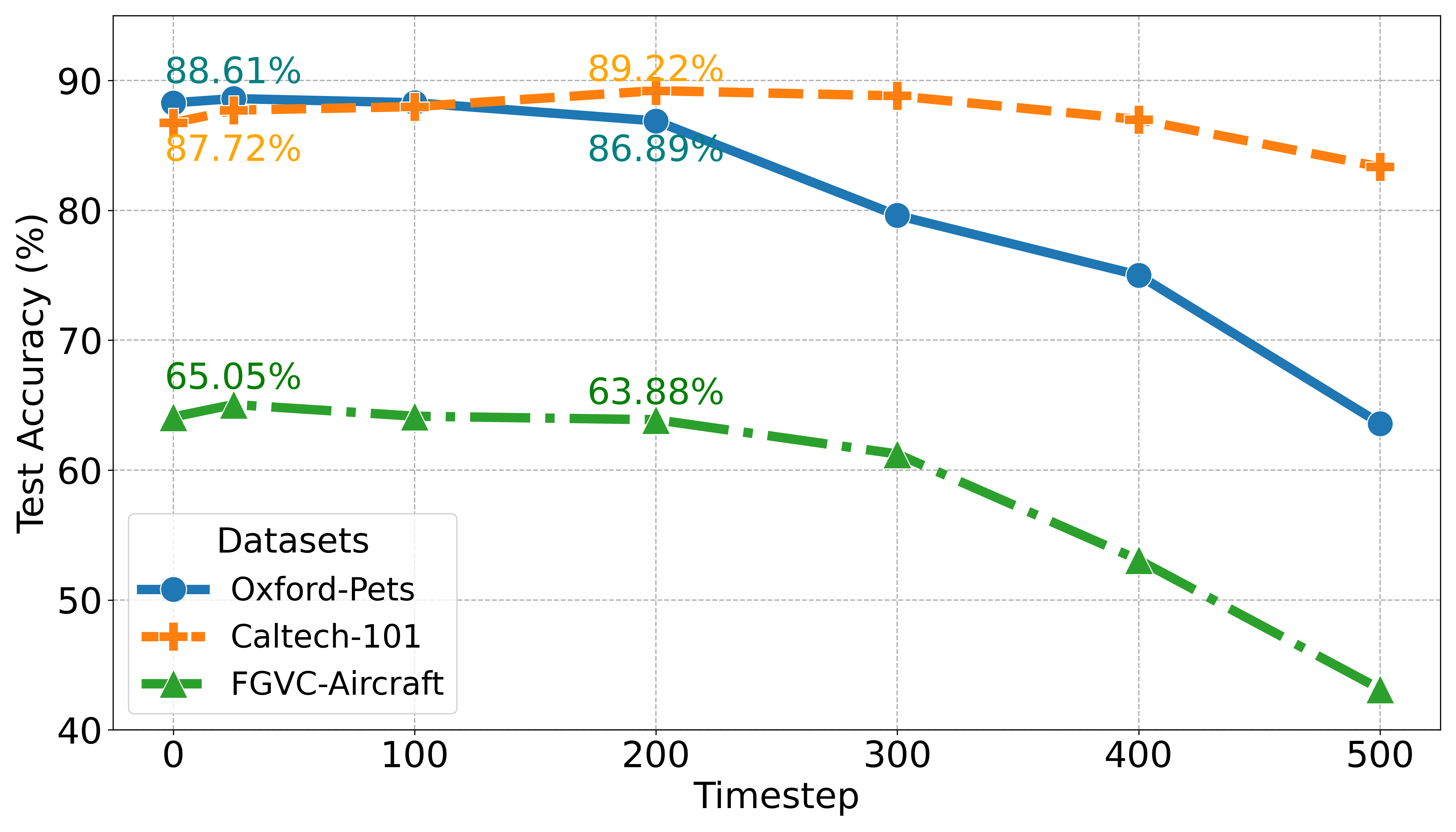}\label{fig:test_acc_decay}}\hfill \\ 
    \vspace{-3pt}
    \caption{\footnotesize \textbf{Top-1 accuracy of \texttt{up\_ft1} at different timesteps:} Earlier timesteps perform better on fine-grained datasets ({\oxford}, {\aircraft}); interim ones on coarse-grained dataset ({\caltech}).}
    \vspace{-0.2in}
    \label{fig:cap_test_acc_decay}
\end{figure}

%% file: sec/expts_architectures.tex
\subsection{Effect of different models and architectures}\label{sec:arc}
\input{fig_latex/step25_up1_SD21}
\input{fig_latex/sd_vs_if}

\input{tables_latex/model_arch_pets_comp}
\input{fig_latex/dit.tex}

Next, we inspect how diffusion models that differ in their underlying architectures, pre-training datasets, attention mechanisms differ in their internal encoding of visual semantic information.

\noindent \textbf{Stable diffusion variants:} We study two stable diffusion models ({\sdone} v/s {\sdtwo}) which primarily differ in the underlying text encoder and pre-trained datasets. We first extract diffusion features from \texttt{bottleneck} and \texttt{up\_ft1} at $t=\{25, 200\}$ on {\oxford} dataset and train k-SAE and \textbf{Diff-C} models. From Table~\ref{tab:model}, we note that {\sigmalabel} is lower for {\sdone} indicating that {\sdone} captures more object-specific information compared to {\sdtwo}. This is qualitatively supported by Fig.~\ref{fig:qual_step25_oxfordpet_up1_up2} and Fig.~\ref{fig:step25_up1_sd21} where k-SAE neurons are being activated by images with clearer object-specific information when using {\sdone} features compared to {\sdtwo}. This is also consistent with \textbf{Diff-C} results in Table~\ref{tab:diffusion_model_test_accuracy}, where \texttt{bottleneck} features of {\sdtwo} are particularly under-performing compared to {\sdone}. Even though there is a sharp performance boost of $\mathbf{13.17\%}$ from using \texttt{up\_ft1} for both architectures ($t=25$), {\sdone} performs better overall across timesteps. Similar behavior was noted for zero-shot classification in~\cite{diffusion_classifier_23} but not well-understood and is a fruitful topic for future research. 

\noindent \textbf{Latent v/s pixel space:} We next examine how diffusion denoising in the pixel space impacts the learnt visual information differently from those learnt in the latent space. To this end, we compare classification performance of stable diffusion features with those from DeepFloyd-IF~\cite{DeepFloydIF} which operates directly in the pixel space\footnote{We acknowledge that despite having similar number of model parameters, both models have different pre-training data, cross-attention connections, and different initial input image resolution.}. Unlike \sdone, we note an uptick in the performance of features from \texttt{up\_ft0} layer of DeepFloyd-IF by $\mathbf{1.74\%}$ for {\oxford}, and $\mathbf{2.85\%}$ for {\caltech} for $t$=25 as illustrated in Fig.~\ref{fig:cap_sd_vs_if}. Furthermore, from Table~\ref{tab:diffusion_model_test_accuracy}, we observe that DeepFloyd-IF’s performance is more sensitive to timesteps than Stable Diffusion. For instance, \sdone\ has a dip of $\mathbf{1.97\%}$ in top-1 accuracy when using \texttt{bottleneck} features at $t=25$ v/s $t=200$, while DeepFloyd-IF has a significant drop of $\mathbf{7.08\%}$. Given DeepFloyd-IF operates directly in the pixel space, we think that each denoising step is introducing larger shifts in the underlying semantic structure than in latent space, contributing to these differences.
\input{fig_latex/main_pca_maps}

\noindent \textbf{Different diffusion architectures:} We also study how semantic information representation varies with the choice of diffusion architecture. To this end, we compare features from U-Net based diffusion model against transformer-based model. Specifically, we extract features from different encoder blocks of DiT~\cite{dit} and interpret them via both k-SAE and \textbf{Diff-C}. From Table~\ref{tab:block}, we observe that the middle block of DiT (\texttt{block14}) yields the lowest {\sigmalabel} compared to earlier and later layers. This is qualitatively supported by Fig.~\ref{fig:dit}, where the \texttt{block14} features contain more class-specific information than other blocks. While with U-Net based features we saw images with spatially related photographic styles emerge (\eg, similar postures or photographic compositions as shown in Fig.~\ref{fig:qual_step25_oxfordpet_up1_up2} (a), (b)), we did not find similar patterns emerge from earlier or later layers of DiT. Though the selected DiT and U-Net based diffusion models have similar number of parameters (Table~\ref{tab:diffusion_model_test_accuracy}), transformer-based DiT may have less spatial inductive biases compared to the convolutional-based U-Net. Additional visualizations of DiT features in suppl. material.

\noindent \textbf{Inductive biases in diffusion models:} To more deeply understand the difference between DiT and {\sdone} in \textit{how} spatial information is internally encoded, we follow the approach from~\cite{plugandplay} and apply principle component analysis (PCA) on the diffusion features, and visualize the first three principal components of images from UnRel~\cite{UnRel} dataset. From Fig.~\ref{fig:pca_unrel_main}, it is evident that \texttt{bottleneck} features of
{\sdone} capture very coarse spatial information, while \texttt{up\_ft1} capture very clear localized semantic information, even on images where common objects occur out of context. As we go deeper into {\sdone}, layer \texttt{up\_ft2} tends to capture more low-level information (more visualizations in suppl. material). By contrast, DiT’s maps exhibit blended colors across all layers, indicating no clear spatially localized information. This property aligns with transformers' tendency to capture more global context by attending to the entire image, and supports k-SAE's interpretations in Fig.~\ref{fig:dit} that there is less spatially-rich information in DiT.

%% file: fig_latex/step25_up1_SD21.tex
\begin{figure}[t]
    \centering
    \vspace{-0.1in}
    {\includegraphics[width=0.7\linewidth]{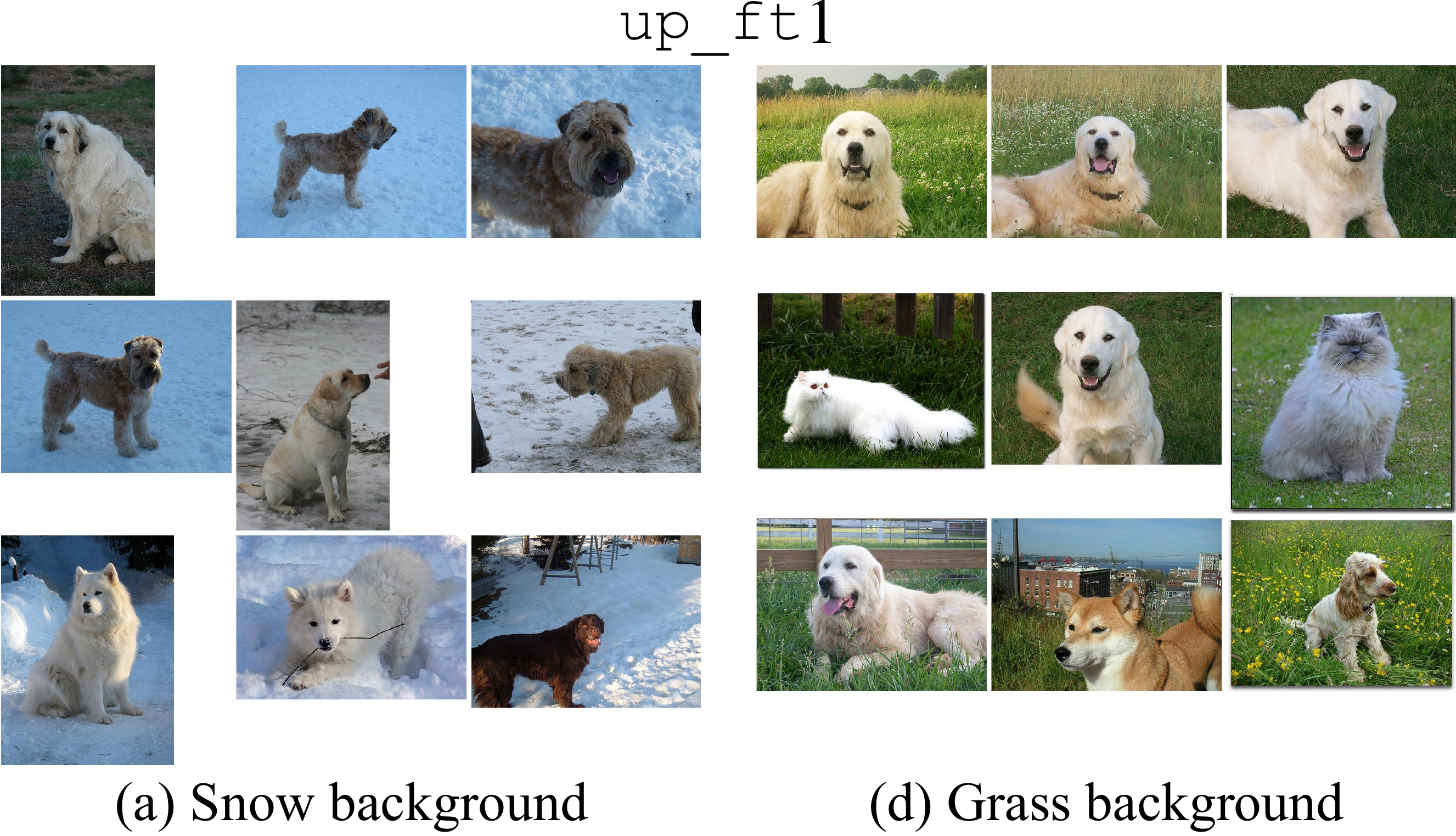}}\hfill \\ 
    \vspace{-2pt}
    \caption{\footnotesize{\textbf{k-SAE visualizations of \texttt{up\_ft1} of {\sdtwo} on \oxford} at $t=25$.
    Contrary to {\sdone} (Fig.~\ref{fig:qual_step25_oxfordpet_up1_up2} (c), (d)) where $8$ out of $9$ images depict the same breed, {\sdtwo} features results in $4$ in $9$ images in (a) as Wheaten Terriers and (b) $5$ in $9$ images are Great Pyrenees in (b). }} 
    \vspace{-0.1in}
    \label{fig:step25_up1_sd21}
\end{figure} 

%% file: fig_latex/sd_vs_if.tex
\begin{figure}[t]
    \centering
    {\includegraphics[width=0.75\linewidth]{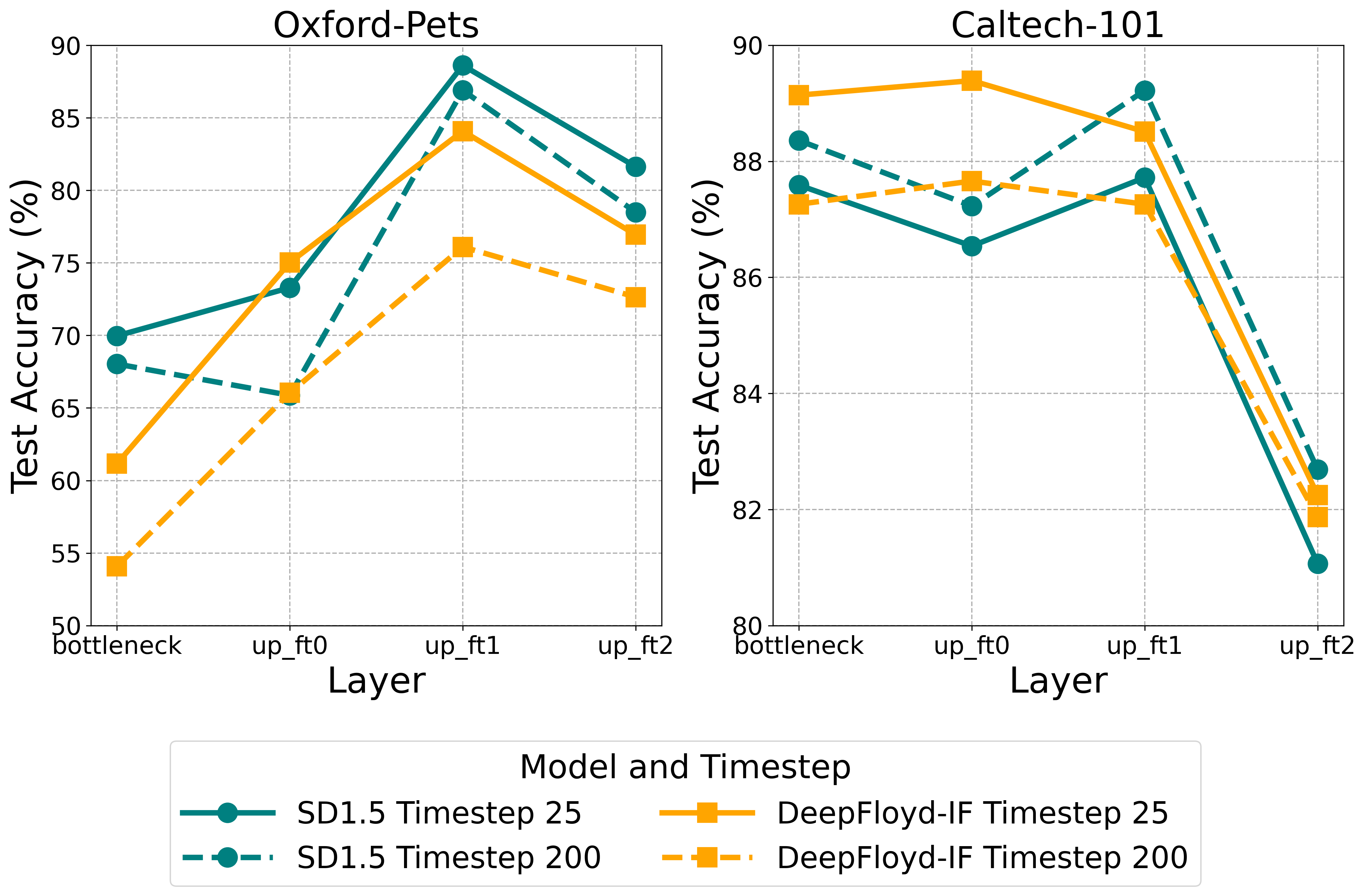}\label{fig:sd_vs_if}}
    \vspace{-0.1in}
    \caption{\footnotesize \textbf{Stable Diffusion vs DeepFloyd-IF:} The performance dip at \texttt{up\_ft0} is not observed for DeepFloyd-IF across both datasets.}\label{fig:cap_sd_vs_if}
    \vspace{-0.1in}
\end{figure}

%% file: tables_latex/model_arch_pets_comp.tex
\begin{table}[t]
    \centering
    \footnotesize
    \setlength{\tabcolsep}{2pt}  
    \renewcommand{\arraystretch}{1}  
    \resizebox{0.35\textwidth}{!}{ 
        \begin{tabular}{ccccc} 
            \toprule
            \textbf{Model} & \textbf{Params (M)} & \textbf{Layer} & $\mathbf{t}$ & \textbf{Test Acc} \\
            \midrule
            \multirow{4}{*}{\sdtwo} & \multirow{4}{*}{900} & \multirow{2}{*}{\texttt{bottleneck}} & 25 & 56.80 \\
                                    &                     &                           & 200 & 56.15 \\
                                    \cmidrule{3-5}
                                    &                     & \multirow{2}{*}{\texttt{up\_ft1}}  & 25 & 84.74 \\
                                    &                     &                           & 200 & 81.77 \\
            \midrule
            \multirow{4}{*}{\sdone} & \multirow{4}{*}{893} & \multirow{2}{*}{\texttt{bottleneck}} & 25 & 69.97 \\
                                    &                     &                           & 200 & 68.03 \\
                                    \cmidrule{3-5}
                                    &                     & \multirow{2}{*}{\texttt{up\_ft1}}  & 25 & \textbf{88.61} \\
                                    &                     &                           & 200 & 86.89 \\
            \midrule
            \multirow{4}{*}{DiT} & \multirow{4}{*}{783} & \multirow{2}{*}{\texttt{block10}} & 25 & 87.49 \\
                                    &                           &                           & 200 & 80.89 \\
                                    \cmidrule{3-5}
                                    &                           & \multirow{2}{*}{\texttt{block14}}  & 25 & \textbf{88.61} \\
                                    &                           &                           & 200 & 83.02 \\
            \midrule
            \multirow{4}{*}{DeepFloyd-IF I-900M} & \multirow{4}{*}{900} & \multirow{2}{*}{\texttt{bottleneck}} & 25 & 61.16 \\
                                    &                           &                           & 200 & 54.08 \\
                                    \cmidrule{3-5}
                                    &                           & \multirow{2}{*}{\texttt{up\_ft1}}  & 25 & 84.08 \\
                                    &                           &                           & 200 & 76.09 \\
            \bottomrule
        \end{tabular}
    }
    \caption{\footnotesize{\textbf{Top-1 accuracy of different diffusion architectures on \oxford.\ }{\sdone}'s \texttt{up\_ft1} and DiT's block 14 perform best overall.}}
    \label{tab:diffusion_model_test_accuracy}
\end{table}

%% file: fig_latex/dit.tex
\begin{figure}[t]
    \centering
    {\includegraphics[width=0.99\linewidth]{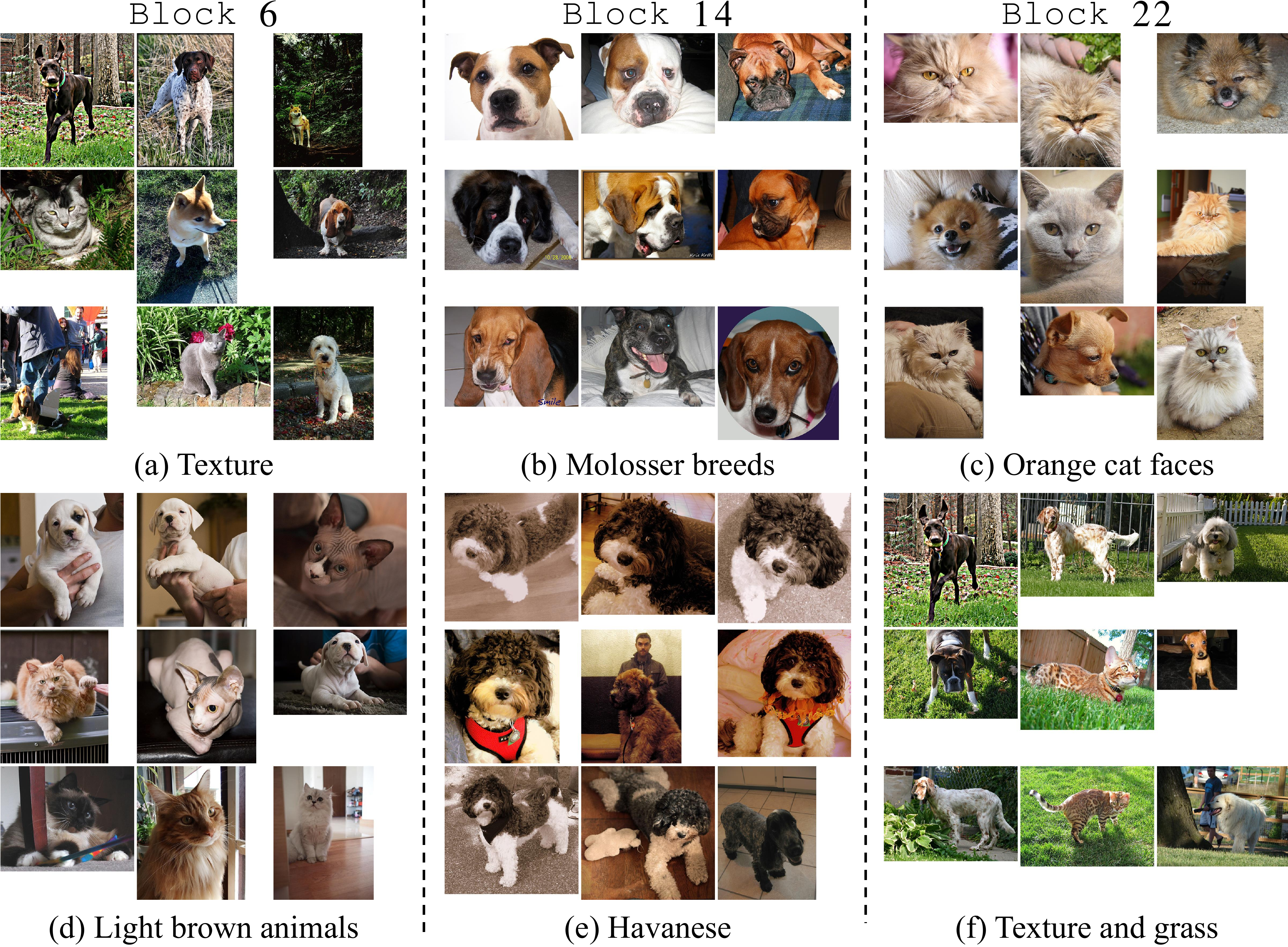}}\hfill \\ 
    \caption{\footnotesize \textbf{k-SAE visualizations of DiT blocks on \oxford}. Block 14 captures fine-grained information; others capture less distinct features.}
    \label{fig:dit}
    \vspace{-0.1in}
\end{figure}

%% file: fig_latex/main_pca_maps.tex
\begin{figure}[t]
    \centering
    \begin{minipage}[b]{0.9\linewidth}
        \centering
        \includegraphics[width=0.8\textwidth]{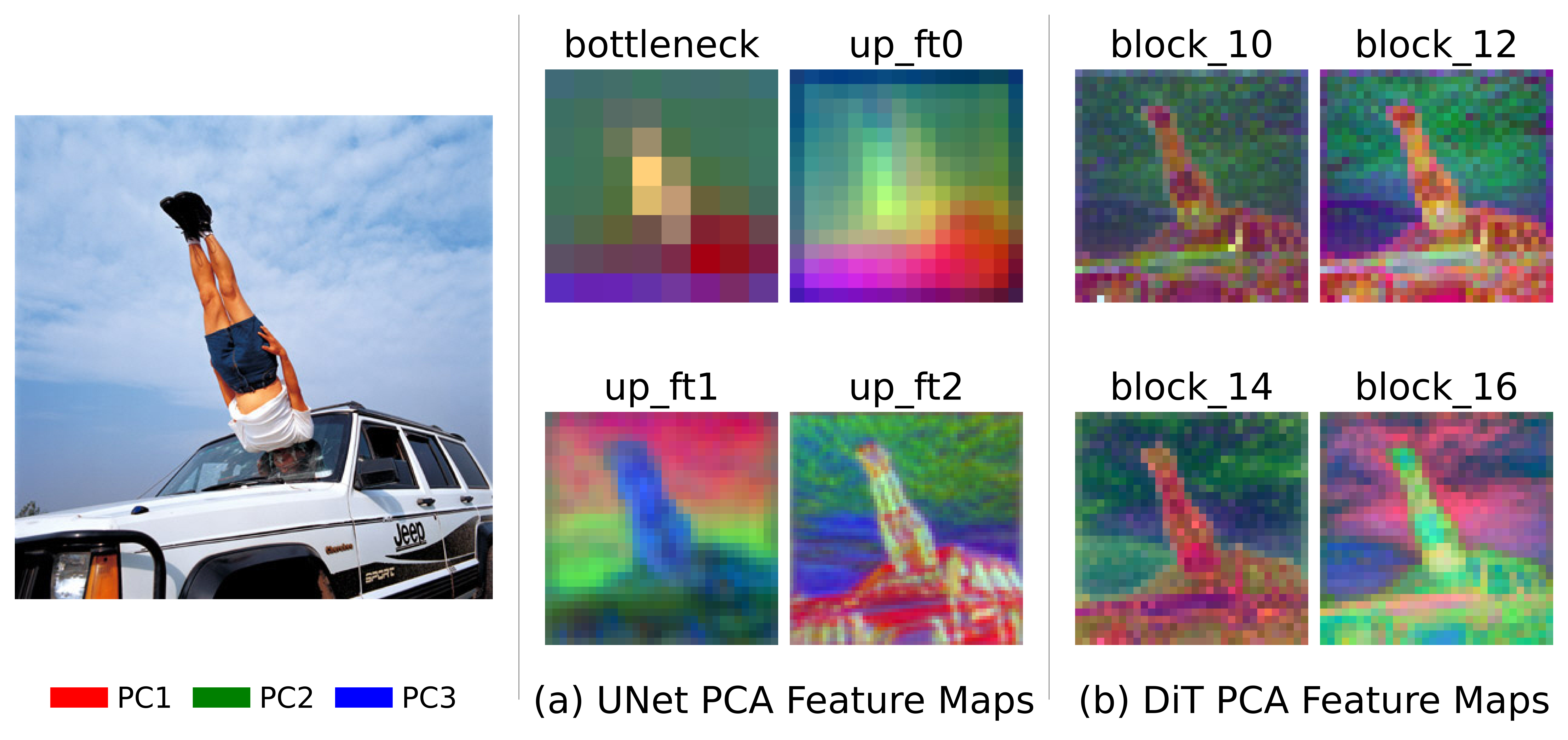}
    \end{minipage}
    \vspace{-0.1in}
    \caption{\footnotesize \textbf{Visualizing top-3 PCA components of diffusion features} from {\sdone} and DiT. \texttt{bottleneck}, \texttt{up\_ft0}, and \texttt{up\_ft1} of {\sdone} capture spatially localized information at varied granularity. This property is missing from DiT features across different blocks.}
    \label{fig:pca_unrel_main}
    \vspace{-0.2in}
\end{figure}

%% file: sec/expts_visualreasoning.tex
\subsection{Performance on visual reasoning}
\label{sec:visual_reasoning}
\input{tables_latex/visual_reas}
Next, we study the generalizability of diffusion features for visual reasoning by integrating them into the LLaVA~\cite{llava} framework. Specifically, on the LLaVA-Lightning configuration~\cite{llava}, we extract features from CLIP~\cite{pmlr-v139-radford21a}, DINO-v2~\cite{Oquab2023DINOv2LR}, and \texttt{up\_ft1} layer from {\sdone} and pass them independently through separate multi-layer projection layers. We then interleave the projected embeddings as done in~\cite{eyeshut} and pass them into the language model in LLaVA~\cite{llava}. 
We use different visual features but keep the language model fixed on the LLaVA-Bench (In-the-Wild)~\cite{llava} evaluation benchmark. We report the relative scores of the model compared to GPT-4 obtained answers aggregated over three categories (complex reasoning, conversational, and descriptive tasks), more details in suppl. material. 

From Table~\ref{tab:llava_scores_main}, it is clear that interleaving CLIP with \texttt{up\_ft1} diffusion features extracted at $t=25$ improves the relative score by $\mathbf{3.3\%}$. By contrast, interleaving CLIP with DINO-v2 features led to a dip in the performance by $\mathbf{9.6\%}$. Additionally, we evaluate LLaVA on MM-Vet benchmark~\cite{yu2023mm} and find that diffusion features yield a correctness score of $26.4$, as assessed by GPT-4o-mini, which is a $\mathbf{+1.1}$ improvement over using only CLIP features. Notably, for the OCR sub task we see a boost of $\textbf{+2.2}$ compared to CLIP features. These results suggest that diffusion features, like CLIP, enjoy the benefit of being multi-modal. However, unlike CLIP, diffusion features also encode strong local semantic information (Fig.~\ref{fig:pca_unrel_main}) making them very powerful feature representations.

%% file: tables_latex/visual_reas.tex
\begin{table}[t]
    \centering
    \setlength{\tabcolsep}{2pt}  
    \renewcommand{\arraystretch}{1.2}  
    \resizebox{0.28\textwidth}{!}{%
        \begin{scriptsize}
            \begin{tabular}{lc}
                \toprule
                \textbf{LLaVA Vision Encoder} & \textbf{Relative Score} \\
                \midrule
                CLIP                 & 56.6 \\
                CLIP + DINO-v2 & 47.0 \\
                CLIP + {\sdone} (\texttt{up\_ft1} at $t=25$)    & \textbf{59.9} \\
                CLIP + {\sdone} (\texttt{up\_ft1} at $t=200$)      & 56.8 \\
                \bottomrule
            \end{tabular}
        \end{scriptsize}%
    }   \vspace{-0.05in}
    \caption{\footnotesize \textbf{Performance of multi-modal reasoning task:} {\sdone}'s \texttt{up\_ft1} features when integrated with CLIP into LLaVA lead to significant boosts on LLaVA-Bench(In-the-Wild), reflecting alignment with the reference answer generated by text-only GPT-4 responses.}
    \label{tab:llava_scores_main}
   \vspace{-0.05in}
\end{table}

%% file: sec/expts_sota.tex
\subsection{State-of-the-art performance}\label{sec:sota}
\input{tables_latex/diffusion_classifier_baselines}
Finally, we compare {\modelName} with other models that use diffusion features for representation learning (Table~\ref{tab:pets_aircraft_baseline_comparison} top row) and also with CLIP variants (Table~\ref{tab:pets_aircraft_baseline_comparison} bottom row). In addition to passing an empty prompt which is our default setting, we also experiment with providing a CLIP-inferred prompt during diffusion feature extraction for a fairer comparison with~\cite{diffusion_classifier_23}. We note that having access to prompts which presumably can have information about the object in the image puts these models at an unfair advantage, but do this only for comparison. From Table~\ref{tab:pets_aircraft_baseline_comparison}, we see that {\modelName} performs significantly better than the best reported numbers in \cite{diffusion_classifier_23}: an improvement of $\mathbf{+1.39\%}$ on {\oxford} ($88.69$) and a huge boost of $\mathbf{+39.03\%}$ for {\aircraft}. ``SD-2.0 features'' baseline from \cite{diffusion_classifier_23} inputs \texttt{bottleneck} features into ResNet~\cite{he2015deepresiduallearningimage} like architecture consuming $520M$ model parameters. On the other hand, {\modelName} is a significantly lighter model ($40M$ model parameters) and yet, achieves a huge boost of $\mathbf{+15.07\%}$ for {\oxford} and $\mathbf{+29.87\%}$ for {\aircraft} from using \texttt{up\_ft1} features. This boost clearly illustrates the effectiveness of interpreting the diffusion model states and making an informed selection for achieving the best transfer learning performance on target tasks. It also highlights the benefit of selecting the right visual features over using complex, highly parameterized models. Crucially, the diffusion classifier from~\cite{diffusion_classifier_23} takes 
$\approx 24$ sec / sample (using their default settings on \oxford) on a single NVIDIA RTX A6000, while {\modelName} takes only $\approx 0.13$ sec / sample, thereby yielding a $4$ orders of magnitude speedup during inference.

\noindent \textbf{Effect of text conditioning:} We note that text-conditioning yields mixed results: it leads to performance improvement of {\modelName} (rows 3 v/s 4 in Table~\ref{tab:pets_aircraft_baseline_comparison}) by $\mathbf{+2.28\%}$ on {\oxford}, but a slight dip of $\mathbf{-0.09\%}$ on {\aircraft} dataset. This detrimental effect of text-conditioning is more pronounced when comparing Diffusion Classifier with SD-2.0 features (rows 1 v/s 2 in Table~\ref{tab:pets_aircraft_baseline_comparison}), where the former uses text information, while SD-2.0 features is based purely on visual features. This behavior is not well understood and could be because of the low frequency of occurrence of specific aircraft model names in the natural language captions used for pre-training or a misalignment between pre-training and the domain-specific aircrafts image data. 

%% file: tables_latex/diffusion_classifier_baselines.tex
\begin{table}[t]
    \centering
    \setlength{\tabcolsep}{2pt}  
    \renewcommand{\arraystretch}{1.2}  
    \resizebox{0.45\textwidth}{!}{%
        \begin{footnotesize}
            \begin{tabular}{lccc}
                \toprule
                \textbf{Model} & \textbf{Num Params ($M$)} & \textbf{\oxford} & \textbf{\aircraft} \\
                \midrule
                Diffusion Classifier (SD-2.0)$^{\dagger}$ \cite{diffusion_classifier_23} & 900 & 87.3* & 26.04* \\
                SD-2.0 features \cite{diffusion_classifier_23} & 1420 & 75.9 & 35.2 \\
                \modelName\ (\texttt{up\_ft1}) - empty & 800 & 88.69 & \textbf{65.07} \\
                \modelName\ (\texttt{up\_ft1}) - from\_CLIP & 800 & 90.97* & 64.98* \\
                \midrule
                \hline
                CLIP ResNet-50 $^{\dagger}$~\cite{pmlr-v139-radford21a} & 102 & 85.4 & 19.3 \\
                OpenCLIP (ViT-H/14)$^{\dagger}$ \cite{openclip} & 630 & \textbf{94.39} & 42.75 \\
                \bottomrule
            \end{tabular}
        \end{footnotesize}%
    }\vspace{-0.05in}
    \caption{\footnotesize \textbf{Top-1 accuracy on {\oxford} and {\aircraft}}. Among models that use diffusion features (top), {\modelName} performs best and competes well with CLIP (bottom) $^{\dagger}$ : zero-shot. *: uses text-conditioning.} \label{tab:pets_aircraft_baseline_comparison}
    \vspace{-0.18in}
\end{table}

%% file: sec/5_conclusion.tex
\section{Discussion and Future work}
In this work, we present k-sparse auto-encoders as an effective tool to dissect diffusion models of different architectures, across different layers, and inference timesteps. Our qualitative and quantitative analysis shows that the abstraction of visual information oscillates from coarse-grained to fine-grained and then back to coarse-grained as we traverse along the depth of a diffusion model. Fruitful research directions entail effective ways to leverage the interpreted information to design better semantic editing algorithms, and dense prediction tasks. Exploring models trained with representation alignment also presents a valuable direction. We hope that our work will spark more interest on the topic of diffusion model interpretability in the research community.

%% file: sec/X_suppl.tex
\clearpage
\maketitlesupplementary

\appendix

\input{sec/expt_backup}

%% file: sec/expt_backup.tex
\input{tables_latex/classification_aircraft_timestep}
\input{tables_latex/classification_pets_timestep}
\input{tables_latex/classification_caltech_timestep}
\input{tables_latex/pets_prompt_comp}
\input{tables_latex/diffusion_classifier_unet}

\section{Text Conditioning in Diffusion Models}
Following the findings from~\cite{diffusion_segmentation}, we report the performance of \textbf{Diff-C} in two text conditioning scenarios: i) empty prompt and ii) a meaningful prompt, e.g., ``a photo of a \{$class\_name$\}, a type of pet'', with the $class\_name$ first inferred through a zero-shot classification with CLIP. The motivation behind reporting both scores is to provide a comprehensive understanding of how text conditioning affects visual features at each layer. We report the classification performance with and without CLIP-inferred captions in Tables~\ref{tab:classification_aircraft_timestep},~\ref{tab:classification_pets_timestep}, and ~\ref{tab:classification_caltech101_timestep}. We note that passing specific class information inferred from CLIP generally helps across all three datasets, layers, and timesteps. To further understand how specific the captions should be, we experiment by passing a generic prompt, e.g., ``a photo of a pet'' during the diffusion process. As shown in Table~\ref{tab:pets_prompt_comp} for \texttt{up\_ft1} layer, on \oxford~\cite{oxford}, compared to the base setting of passing in an empty prompt, using a generic prompt leads to a performance drop by $3.14\%$. This indicates that the specificity of the text being used to condition directly impacts feature representation quality, where more targeted prompts align better with class-relevant features, thereby improving model accuracy. Consequently, using precise text conditioning can lead to considerable gains in performance, particularly in distinguishing nuanced categories. However, this may not always be the case as described in Sec.~\ref{sec:sota}, where for \aircraft~\cite{Maji2013FineGrainedVC} conditioning with the class names led to a dip in classification performance. \\

\input{fig_latex/unrel_pca_maps}
\section{Layer-wise PCA Analysis of Feature Maps}
Figures~\ref{fig:pca_unrel} and~\ref{fig:pca_unrel_dit} provides more evidence to the findings in Sec.~\ref{sec:arc}. by highlighting differences in how {\sdone} and DiT encode spatial information. In {\sdone}, the feature maps reveal well-defined spatial structures, with consistent colors and textures that correspond to specific regions in the image. By contrast, the feature maps of DiT display blended patterns, suggesting a stronger focus on capturing global context rather than emphasizing distinct spatial details.

\input{tables_latex/visual_reas_suppl}
\section{Additional Details on the Visual Reasoning Task}
\input{tables_latex/llava_setup}

\textbf{Hyper-parameters:} We adopt the same hyperparameters used in the the  LLaVA-Lightning~\cite{llava} configuration across all experiments. We use MPT-7B-Chat~\cite{MosaicML2023Introducing} as the language model, and CLIP ViT-L/14~\cite{pmlr-v139-radford21a}, DINOv2 ViT-L/14~\cite{Oquab2023DINOv2LR}, {\sdone} as the vision encoders.  We show the training hyperparameters in Table~\ref{tab:llava_hyperparameters}. All experiments were conducted using a maximum of 4  NVIDIA RTX A6000 GPUs.\\

\noindent \textbf{Pre-training datasets:} Following LLaVA-Lightning~\cite{llava}, we use CC595k~\cite{sharma-etal-2018-conceptual} for stage 1 pre-training, to align the visual encoder with the language model to establish a shared vision-language representation, by tuning the adapter. For stage 2 fine-tuning we use LLaVA-Instruct-80K~\cite{llava} to fine-tune the model to enhance instruction-following capabilities. \\

\noindent \textbf{Adapter settings:} For experiments involving CLIP and DINOv2 features, we use the standard 2 layer MLP projector to align visual tokens with language tokens~\cite{llava}. To obtain tokenized representations from the feature maps obtained from {\sdone}, we first add a 2 layer convolutional block and transform the feature map into pseudo-tokenized representations that match the token embedding dimensions of CLIP and DINOv2. These pseudo-tokenized representations are then passed into the 2 layer MLP projector for alignment.\\

\noindent \textbf{Interleaving diffusion features with CLIP for visual reasoning tasks:} For the experiments reported in  Sec.~\ref{sec:visual_reasoning}, we first gradually reduce the spatial dimension of \texttt{up\_ft1} from  $1280 \times 32 \times 32$ to $256 \times 1024$ to match the token dimensions of CLIP vision embeddings. Next, we process these embeddings through two separate multi-layer projection layers resulting in projected embeddings of shape $256 \times 4096$. Finally, we interleave the projected token embeddings as done in~\cite{eyeshut} before passing them into LLaVA~\cite{llava}. \\

\noindent \textbf{Performance:} Table~\ref{tab:llava_all_scores} compares the performance of different vision encoders in LLaVA, including CLIP (Table~\ref{tab:llava_a}), CLIP+DINOv2 (Table~\ref{tab:llava_b}), and CLIP+Diffusion at timesteps $t=25$ (Table~\ref{tab:llava_c}) and $t=200$ (Table~\ref{tab:llava_d}). The evaluation is conducted on the LLaVA-Bench (in-the-wild)~\cite{llava} benchmark. The benchmark evaluates models across four categories: overall performance (`all'), complex reasoning (`LLaVA Bench complex'), conversational tasks (`LLaVA Bench conversational'), and descriptive tasks (`LLaVA Bench detail').

For the `detail' category, CLIP+Diffusion at $t=25$ achieves the highest relative score of \textbf{\boldmath $56.2$}, outperforming both CLIP ($50.4$) and CLIP+DINOv2 ($37.7$). This demonstrates that the interleaved diffusion and CLIP features effectively capture fine-grained visual details. In the `complex' category, CLIP+Diffusion at $t=200$ achieves the highest relative score of \textbf{\boldmath $70.5$}, surpassing CLIP ($68.4$). At $t=25$, CLIP+Diffusion scores $67.9$ indicating that the coarser-grained features extracted at higher timesteps ($t=200$) seem more effective for this specific task that requires broader contextual understanding. Next, for the `conversational' category, CLIP+Diffusion at $t=25$ achieves a relative score of \textbf{\boldmath $51.3$}, outperforming both CLIP ($43.9$) and CLIP+DINOv2 ($35.6$). The interleaving of diffusion and CLIP features significantly enhances the model's ability to handle visually grounded conversational tasks effectively.

Finally, we report the overall performance under the `all' category and note that CLIP+Diffusion achieves a superior performance with a score of \textbf{\boldmath $59.9$} at $t=25$, outperforming CLIP’s standalone score of $56.6$. This reinforces the power of the visual representations learnt from the diffusion process in achieving top-performance on diverse vision-language tasks.\\

\section{Additional k-SAE Visualizations}
\noindent \textbf{DiT vs U-Net:} In this section, we provide additional visualizations of k-SAE features. As shown in Fig.~\ref{fig:dit_supp} (b), (e), \texttt{Block 14} of DiT captures more class-specific information than other blocks which is qualitatively corroborated in Table~\ref{tab:block}. However, compared to \sdone, DiT captures less distinct class information, as seen in the snow background in Fig.~\ref{fig:dit_supp} (h). Moreover, the spatially related photographic styles observed in Sec.~\ref{sec:arc} do not emerge in DiT. We hypothesize that the transformer-based relies less on inductive bias information compared to UNet-based \sdone, as discussed in Sec.~\ref{sec:arc}.\\

\noindent \textbf{Later timesteps:} Figure~\ref{fig:step500_up1} presents k-SAE visualization at $t=500$ for \sdone. Compared to $t=25$, features at $t=500$ focus more on low-level information, such as texture and low-light, which is qualitatively corroborated in Table~\ref{tab:timestep}. We hypothesize that as the diffusion timestep increases, so does the added noise, rendering the features less useful for transfer learning, consistent with our observations in Sec.~\ref{sec:timesteps}.

\input{fig_latex/dit_supp}
\input{fig_latex/step200_up1}

\input{tables_latex/ksae_rest}

\section{Additional k-SAE Experiments}
\noindent\textbf{Effect of image resolutions:} To assess the robustness of our method across varying input resolutions, we conduct additional experiments examining the effect of image resolution. As shown in Table~\ref{tab:res}, using different image resolutions exhibits a similar trend in terms of \sigmalabel, with smaller resolutions resulting in slightly reduced variance across different DiT blocks on \oxford.

\section{Additional Implementation Details}
In this section, we provide additional implementation details for training k-SAE.
We set the expansion factor for the k-SAE to $64$, following prior work~\cite{fry2024towards}, resulting in $n = 1280\times64$ = $81,920$ latents for SD and $n = 1152 \times 64$ = $73,728$ latents for DiT. We apply a unit normalization constraint~\cite{sharkey2023taking} on the decoder weights $W_{dec}$ of the k-SAE after each update. We use the Adam~\cite{kingma2014adam} optimizer with a learning rate of $0.0004$ and apply a constant warm up for $500$ steps. The total training time is approximately $1$ hour with $\sim$18 GB peak memory on $1$ NVIDIA RTX A6000 GPU trained for $10M$ steps.

\section{Additional Details of Evaluation}
{In this section, we provide additional details on how we quantify the granularity of semantic information in diffusion features through a multiple-choice question-answering task, as discussed in Sec.~\ref{sec:layers}.
Using GPT-4o~\cite{gpt4o}, we evaluate the level of semantic detail captured by different diffusion features. 
Table~\ref{tab:prompt} presents the prompt used to query the model for this evaluation.
Specifically, we assess the model's predictions based on the top 10 most highly activating images among the top 100 most highly activating neurons of the learned k-SAE.
}

\input{tables_latex/chatgpt_prompt}

%% file: tables_latex/classification_aircraft_timestep.tex
\begin{table*}[t!]
    \centering
    \small
    \setlength{\tabcolsep}{4pt} 
    \renewcommand{\arraystretch}{1.2}
    \resizebox{\textwidth}{!}{
        \begin{tabular}{ccccc}
            \toprule
            \textbf{Timestep (t)} & \textbf{bottleneck (empty / from\_CLIP)} & \textbf{up\_ft0 (empty / from\_CLIP)} & \textbf{up\_ft1 (empty / from\_CLIP)} & \textbf{up\_ft2 (empty / from\_CLIP)} \\
            \midrule
            0   & 52.27 / 52.74 & 48.79 / 49.06 & 64.09 / 62.826   & 50.55 / 49.15 \\
            25  & 51.76 / 54.88 & 50.49 / 51.19 & 65.07  / 63.69  & 50.25 / 49.21 \\
            100 & 51.07 / 55.12 & 49.48 / 51.10 & 64.15  / 64.98   & 51.37 / 50.53  \\
            200 & 50.91 / 52.51 & 49.63 / 49.99 & 63.88  / 63.13   & 50.53 / 50.55 \\
            \bottomrule
        \end{tabular}
    }
    \caption{\textbf{Top-1 accuracy} at different timesteps and layers for fine-grained task (\aircraft).}
    \label{tab:classification_aircraft_timestep}
\end{table*}

%% file: tables_latex/classification_pets_timestep.tex
\begin{table*}[t!]
    \centering
     \footnotesize
    \setlength{\tabcolsep}{4pt} 
    \renewcommand{\arraystretch}{1.2}
    \resizebox{\textwidth}{!}{
        \begin{tabular}{ccccc}
            \toprule
            \textbf{Timestep (t)} & \textbf{bottleneck (empty / from\_CLIP)} & \textbf{up\_ft0 (empty / from\_CLIP)} & \textbf{up\_ft1 (empty / from\_CLIP)} & \textbf{up\_ft2 (empty / from\_CLIP)} \\
            \midrule
            0   & 68.79 / 84.33 & 66.99 / 79.50 & 88.28 / 90.11   & 77.79 / 80.67 \\
            25  & 69.97 / 85.25 & 73.29 / 84.17 & 88.61  / 90.68  & 81.63 / 85.28 \\
            100 & 69.53 / 85.88 & 67.07 / 81.33 & 88.29  / 90.97   & 78.82 / 84.36  \\
            200 & 68.03 / 86.43 & 65.87 / 81.79 & 86.89  / 90.32   & 78.49 / 84.79 \\
            \bottomrule
        \end{tabular}
    }
    \caption{\footnotesize \textbf{Top-1 accuracy} at different timesteps and layers for fine-grained task (\oxford).}
    \label{tab:classification_pets_timestep}
\end{table*}

%% file: tables_latex/classification_caltech_timestep.tex
\begin{table*}[t!]
    \centering
     \footnotesize
    \setlength{\tabcolsep}{4pt} 
    \renewcommand{\arraystretch}{1.2}
    \resizebox{\textwidth}{!}{
        \begin{tabular}{ccccc}
            \toprule
            \textbf{Timestep (t)} & \textbf{bottleneck (empty / from\_CLIP)} & \textbf{up\_ft0 (empty / from\_CLIP)} & \textbf{up\_ft1 (empty / from\_CLIP)} & \textbf{up\_ft2 (empty / from\_CLIP)} \\
            \midrule
            0   & 85.83 / 92.13  & 85.75 / 89.68  & 86.75 / 88.18 & 79.11 / 80.28 \\
            25  & 87.59 / 91.32 & 86.54 / 90.81 & 87.72 / 91.08 & 81.07 / 82.69 \\
            100 & 88.28 / 92.41 & 88.18 / 90.66 & 87.99 / 92.05 & 82.02 / 84.73 \\
            200 & 88.36 / 91.65 & 87.23 / 90.73 & 89.22 / 92.26 & 82.69 / 85.63 \\
            \bottomrule
        \end{tabular}
    }
    \caption{\footnotesize \textbf{Top-1 accuracy} at different timesteps for coarse-grained task (\caltech).}
    \vspace{-0.2in}
    \label{tab:classification_caltech101_timestep}
\end{table*}

%% file: tables_latex/pets_prompt_comp.tex
\begin{table}[h]
    \centering
    \small
    \setlength{\tabcolsep}{6pt} 
    \renewcommand{\arraystretch}{1.2}
    \begin{tabular}{ccccc}
        \toprule
        \multirow{2}{*}{Prompt Type} & \multicolumn{2}{c}{Timestep} \\
        \cmidrule(lr){2-3}
         & 25 & 200 \\
        \midrule
        Empty Prompt & 88.61 & 86.89 \\
        from\_CLIP & \increase{2.34} & \increase{3.94} \\
        generic  & \decrease{3.14} & \decrease{3.13} \\
        \bottomrule
    \end{tabular}
    \caption{\footnotesize \textbf{Performance vs Text Conditioning on \oxford\ using \texttt{up\_ft1}:} Using a generic prompt (``A photo of a pet'') leads to a dip in classification performance compared to using an empty prompt. By contrast, using a targeted caption (``A photo of a \{$class\_name$\}, a type of pet'') leads to a boost in performance.}
    \label{tab:pets_prompt_comp}
\end{table}

%% file: tables_latex/diffusion_classifier_unet.tex
\begin{table}[t]
    \centering
    \scriptsize
    \resizebox{0.35\textwidth}{!}{%
        \begin{tabular}{|l|c|c|}
            \hline
            \textbf{Layer} & \textbf{Output Shape} & \textbf{Description} \\
            \hline
            conv1 & $[B, 1024, H, W]$ & conv2D, $3 \times 3$ \\
            \hline
            conv2 & $[B, 1024, H/2, W/2]$ & conv2D, $3 \times 3$, stride 2 \\
            \hline
            conv3 & $[B, 1024, H/4, W/4]$ & conv2D, $3 \times 3$, stride 2 \\
            \hline
            conv4 & $[B, 1024, H/8, W/8]$ & conv2D, $3 \times 3$, stride 2 \\
            \hline
            GAP & $[B, 1024, 1, 1]$ & global average pooling \\
            \hline
            FC & $[B, \texttt{NUM\_CLASSES}]$ & flatten + FC layer \\
            \hline
        \end{tabular}%
    }
    \vspace{-0.1in}
   \caption{\footnotesize{Architecture of \textbf{\modelName} (\textnormal{$40M$ params})}.}
   \vspace{-0.2in}
    \label{tab:diffc_unet}
\end{table}

%% file: fig_latex/unrel_pca_maps.tex
\begin{figure*}[h]
    \centering
    \begin{minipage}[b]{0.45\textwidth}
        \centering
        \includegraphics[width=\textwidth]{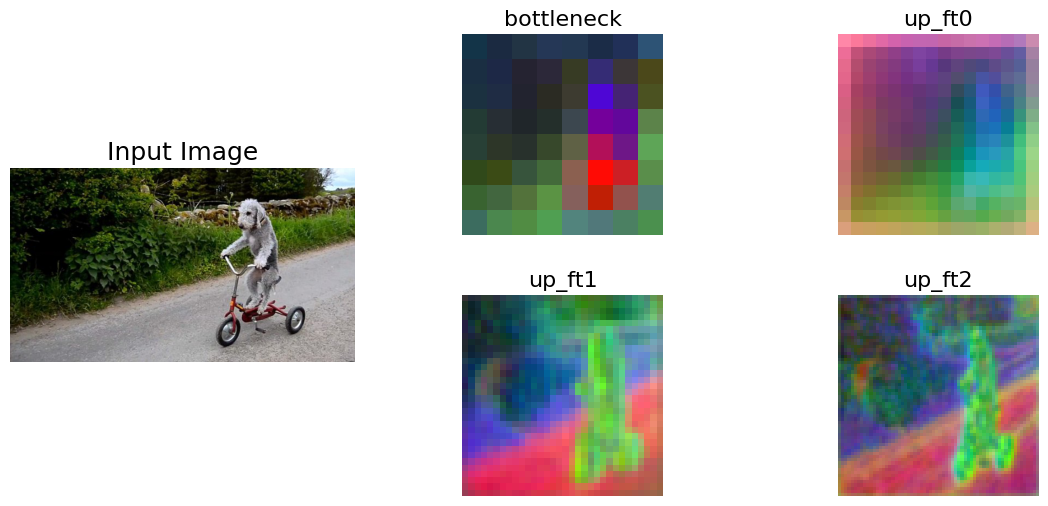}
    \end{minipage}
    \hfill
    \begin{minipage}[b]{0.45\textwidth}
        \centering
        \includegraphics[width=\textwidth]{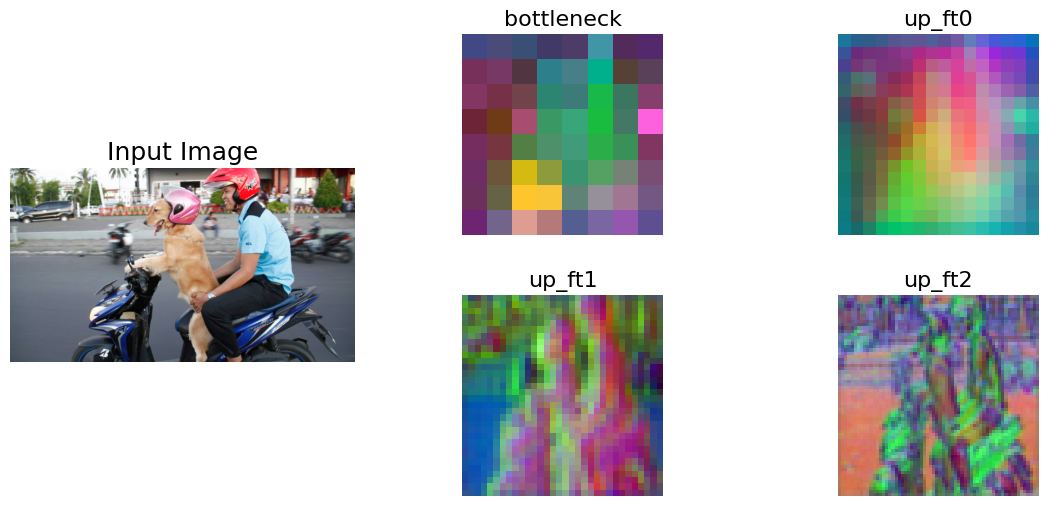}
    \end{minipage}
    
    \vspace{0.5cm}
    
    \begin{minipage}[b]{0.45\textwidth}
        \centering
        \includegraphics[width=\textwidth]{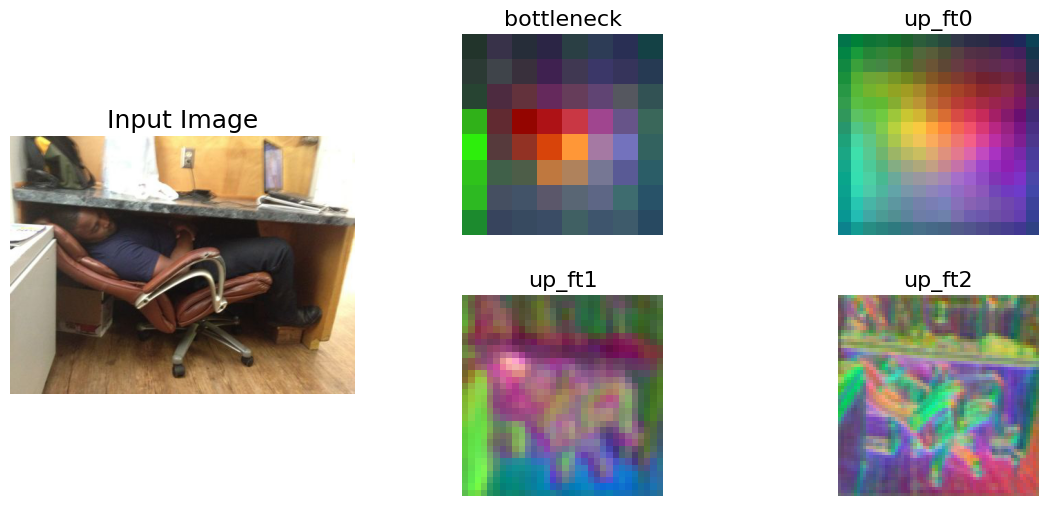}
    \end{minipage}
    \hfill
    \begin{minipage}[b]{0.45\textwidth}
        \centering
        \includegraphics[width=\textwidth]{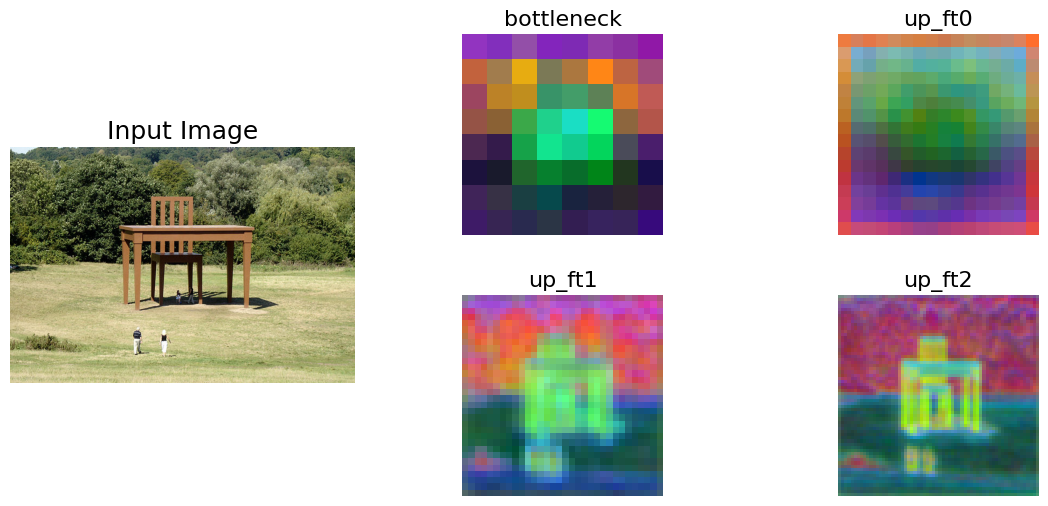}
    \end{minipage}
    \caption{\footnotesize \textbf{PCA Feature Maps SD-1.5 on images} from UnRel~\cite{UnRel} - Consistency of colors and textures (at up\_ft1, up\_ft2) suggests that the model preserves local details and spatial relationships}
    \label{fig:pca_unrel}
\end{figure*}

\begin{figure*}[t]
    \centering
    \begin{minipage}[b]{0.45\textwidth}
        \centering
        \includegraphics[width=\textwidth]{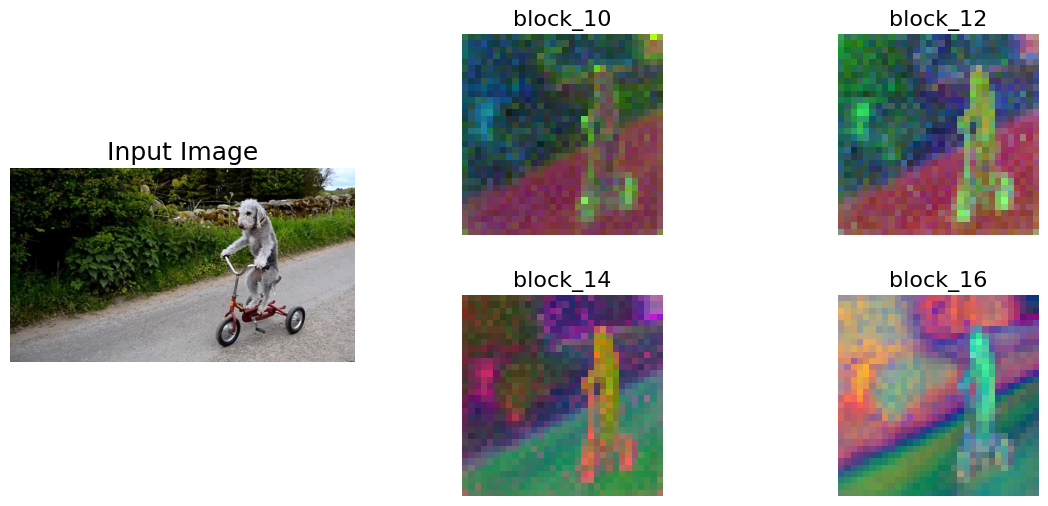}
    \end{minipage}
    \hfill
    \begin{minipage}[b]{0.45\textwidth}
        \centering
        \includegraphics[width=\textwidth]{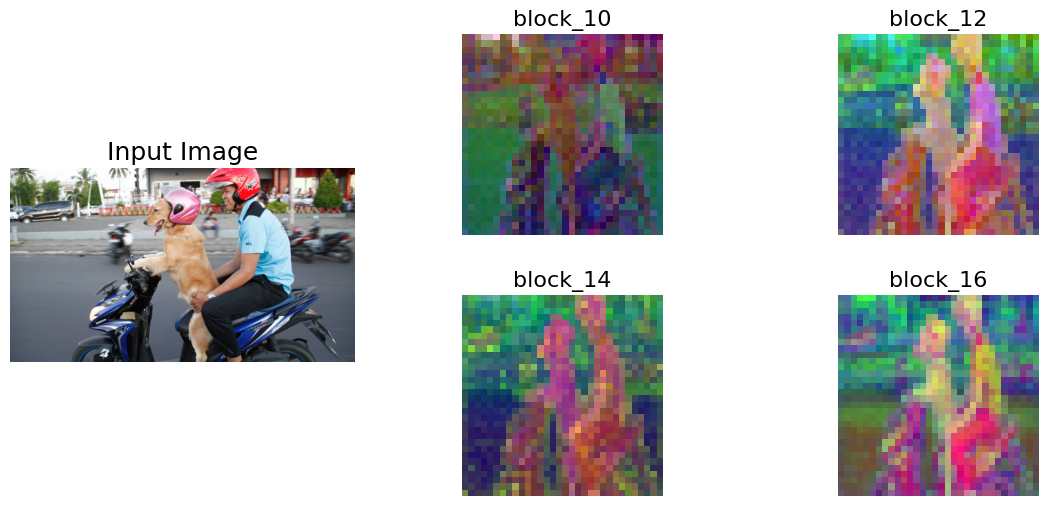}
    \end{minipage}
    
    \vspace{0.5cm}
    
    \begin{minipage}[b]{0.45\textwidth}
        \centering
        \includegraphics[width=\textwidth]{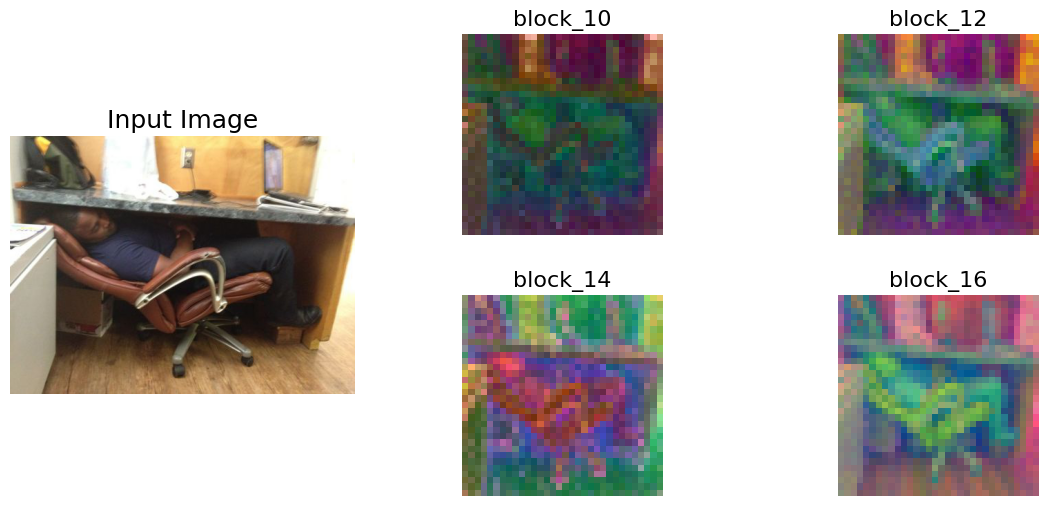}
    \end{minipage}
    \hfill
    \begin{minipage}[b]{0.45\textwidth}
        \centering
        \includegraphics[width=\textwidth]{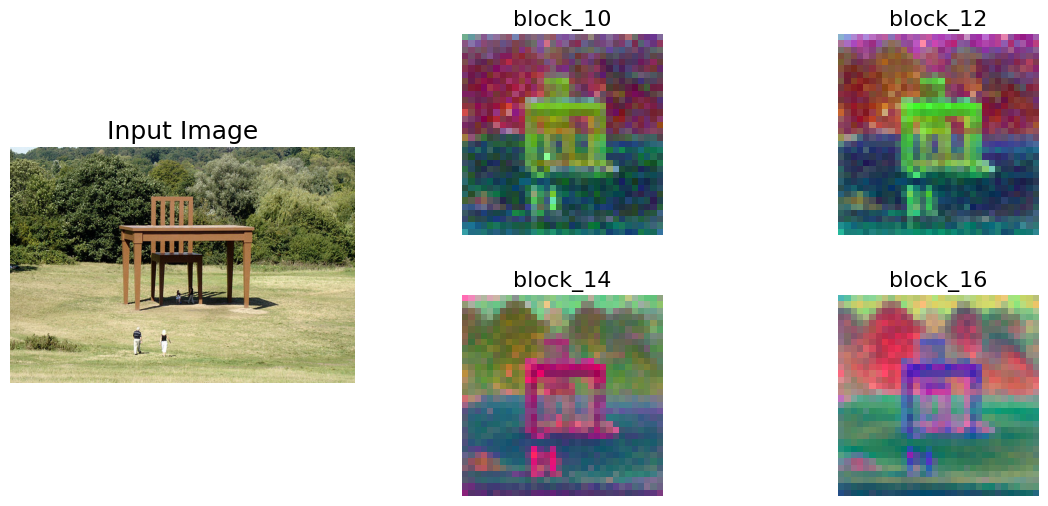}
    \end{minipage}
    \caption{\footnotesize \textbf{PCA Feature Maps DiT on images from UnRe}l~\cite{UnRel} - The blending of colors suggests that the model encodes global relationships while maintaining a holistic representation of spatial structures, rather than isolating precise local details.}
    \label{fig:pca_unrel_dit}
\end{figure*}

%% file: tables_latex/visual_reas_suppl.tex
\begin{table}[t]
    \centering
    \renewcommand{\arraystretch}{1.2}  
    \setlength{\tabcolsep}{6pt}        

    \begin{subtable}[t]{0.48\textwidth}
        \centering
        \resizebox{\textwidth}{!}{%
            \begin{tabular}{lccc}
                \toprule
                \textbf{Category} & \textbf{Relative Score} & \textbf{GPT-4 Score} & \textbf{LLaVA Score} \\
                \midrule
                All                 & 56.6         & 82.7                   & 46.8                    \\
                LLaVA Bench complex & 68.4         & 80.4                   & 55.0                    \\
                LLaVA Bench conversational    & 43.9         & 87.1                   & 38.2                    \\
                LLaVA Bench detail  & 50.4         & 82.0                   & 41.3                    \\
                \bottomrule
            \end{tabular}
        }
        \caption{\footnotesize CLIP LLaVA}
        \label{tab:llava_a}
    \end{subtable}%

    \vspace{12pt}
    
    \begin{subtable}[t]{0.48\textwidth}
        \centering
        \resizebox{\textwidth}{!}{%
            \begin{tabular}{lccc}
                \toprule
                \textbf{Category} & \textbf{Relative Score} & \textbf{GPT-4 Score} & \textbf{LLaVA Score} \\
                \midrule
                All                 & 47.0         & 84.8                   & 39.8                    \\
                LLaVA Bench complex & 59.9         & 81.1                   & 48.6                    \\
                LLaVA Bench conversational    & 35.6         & 94.1                   & 33.5                    \\
                LLaVA Bench detail  & 37.7         & 81.3                   & 30.7                    \\
                \bottomrule
            \end{tabular}
        }
        \caption{\footnotesize CLIP+DINOv2 LLaVA}
        \label{tab:llava_b}
    \end{subtable}
    
    \vspace{12pt}  

    \begin{subtable}[t]{0.48\textwidth}
        \centering
        \resizebox{\textwidth}{!}{%
            \begin{tabular}{lccc}
                \toprule
                \textbf{Category} & \textbf{Relative Score} & \textbf{GPT-4 Score} & \textbf{LLaVA Score} \\
                \midrule
                All                 & \textbf{59.9}         & 83.2                   & 49.8                    \\
                LLaVA Bench complex & 67.9         & 80.0                   & 54.3                    \\
                LLaVA Bench conversational    & \textbf{51.3}         & 90.6                   & 46.5                    \\
                LLaVA Bench detail  & \textbf{56.2}         & 80.7                   & 45.3                    \\
                \bottomrule
            \end{tabular}
        }
        \caption{\footnotesize CLIP+Diffusion ($t=25$) LLaVA}
        \label{tab:llava_c}
    \end{subtable}%

    \vspace{12pt}
    
    \begin{subtable}[t]{0.48\textwidth}
        \centering
        \resizebox{\textwidth}{!}{%
            \begin{tabular}{lccc}
                \toprule
                \textbf{Category} & \textbf{Relative Score} & \textbf{GPT-4 Score} & \textbf{LLaVA Score} \\
                \midrule
                All                 & 56.8         & 83.7                   & 47.5                    \\
                LLaVA Bench complex & \textbf{70.5}         & 80.0                   & 56.4                    \\
                LLaVA Bench conversational    & 45.6         & 87.7                   & 40.0                    \\
                LLaVA Bench detail  & 45.7         & 86.0                   & 39.3                    \\
                \bottomrule
            \end{tabular}
        }
        \caption{\footnotesize CLIP+Diffusion ($t=200$) LLaVA}
        \label{tab:llava_d}
    \end{subtable}
    \caption{\footnotesize \textbf{Performance on the multi-modal reasoning task for various LLaVA configurations}. The integration of Diffusion features with CLIP improves performance across all tasks, with notable gains in the `detail' and `conversational' categories.}
    \label{tab:llava_all_scores}
\end{table}

%% file: tables_latex/llava_setup.tex
\begin{table}[ht]
    \centering
    \small
    \setlength{\tabcolsep}{6pt} 
    \renewcommand{\arraystretch}{1.2}
    \begin{tabular}{lcc}
        \toprule
        \multirow{2}{*}{Hyperparameter} & \multicolumn{2}{c}{Stage} \\
        \cmidrule(lr){2-3}
         & Stage 1 & Stage 2 \\
        \midrule
        batch size & 128 & 128 \\
        learning rate (lr) & 2e-3 & 2e-5 \\
        lr schedule decay & cosine & cosine \\
        lr warmup ratio & 0.03 & 0.03 \\
        weight decay & 0 & 0 \\
        epoch & 1 & 1 \\
        optimizer & \multicolumn{2}{c}{AdamW~\cite{adamw}} \\
        deepspeed stage & 2 & 3 \\
        \bottomrule
    \end{tabular}
    \caption{\footnotesize Hyperparameters for LLaVA-Lightning (default setting)}
    \label{tab:llava_hyperparameters}
\end{table}

%% file: fig_latex/dit_supp.tex
\begin{figure}[t]
    \centering
    {\includegraphics[width=0.95\linewidth]{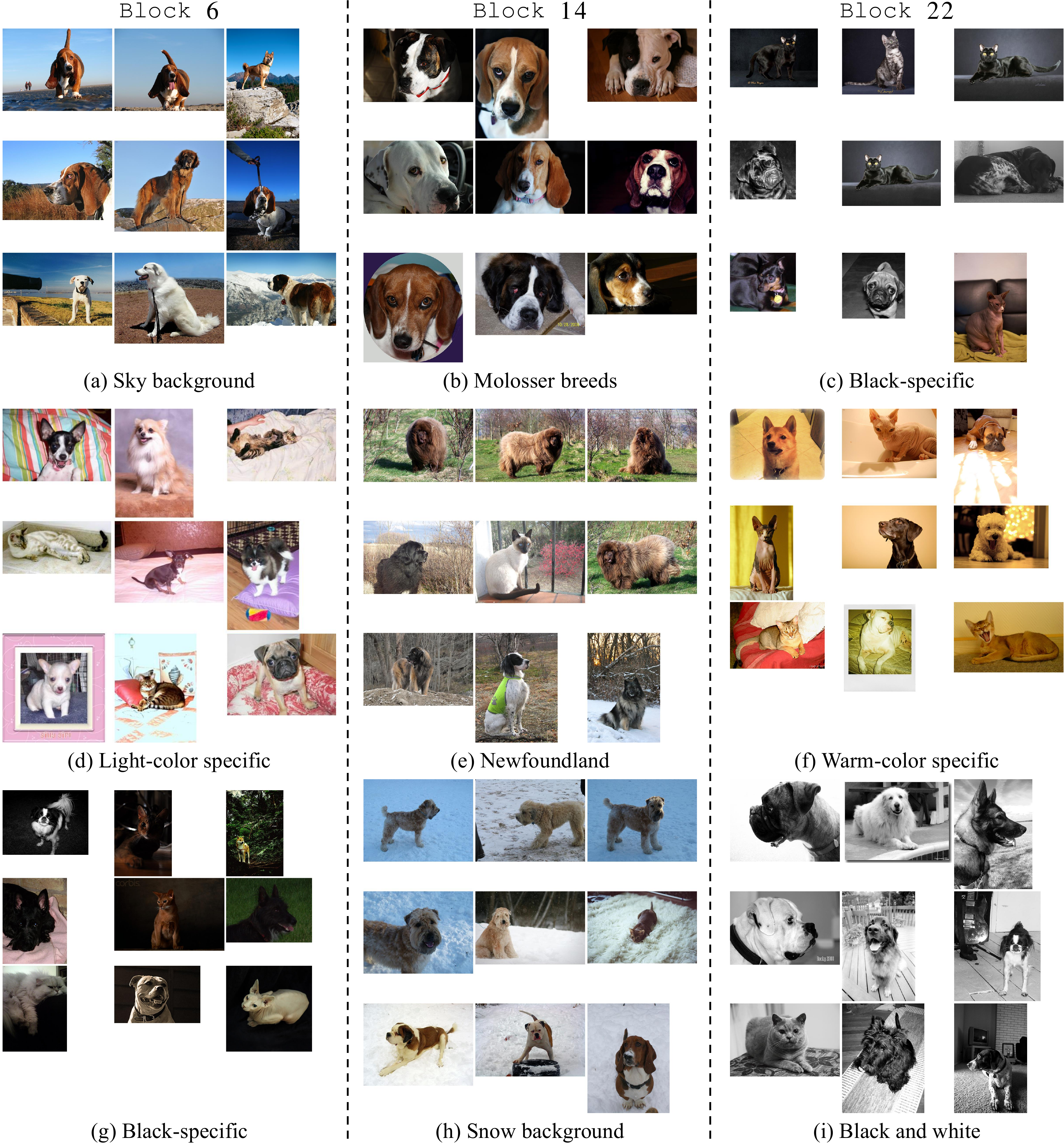}}\hfill \\ 
    \caption{\footnotesize \textbf{k-SAE visualizations of the blocks on \oxford} at $t = 25$. 
    \texttt{Block 14} mainly captures class-specific information, while other blocks focus more on less distinct features. }
    \label{fig:dit_supp} 
\end{figure} 

%% file: fig_latex/step200_up1.tex
\begin{figure}[t]
    \centering
    {\includegraphics[width=0.7\linewidth]{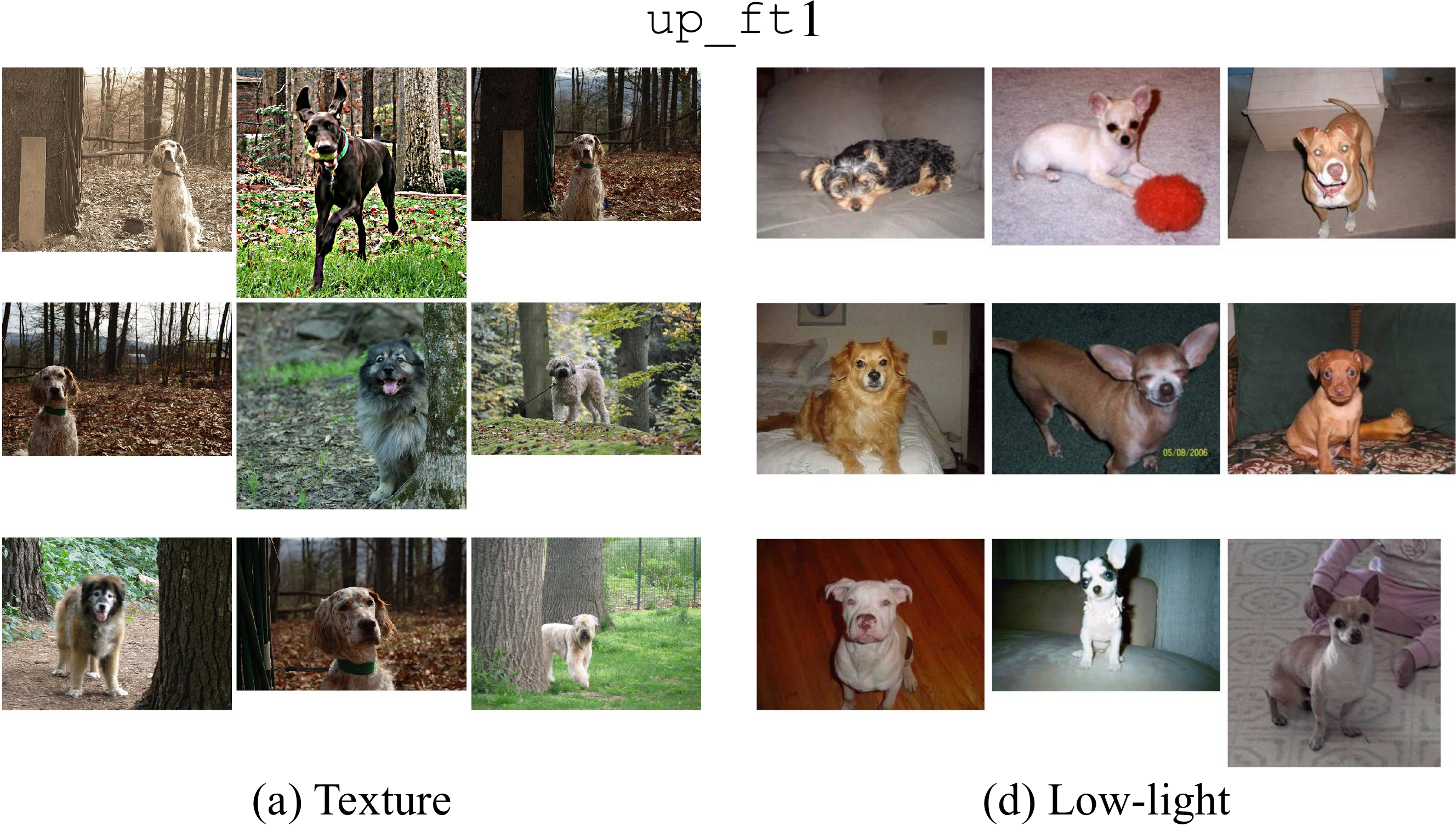}}\hfill \\ 
    \caption{\footnotesize \textbf{k-SAE visualizations on \oxford} of \texttt{up\_ft1} UNet layer at $t=500$. 
    In contrast to the earlier timestep (Fig~\ref{fig:qual_step25_oxfordpet_up1_up2}), $t=500$ appears to focus more on low-level features.}
    \label{fig:step500_up1}
\end{figure}

%% file: tables_latex/ksae_rest.tex
\begin{table}[t]
    \centering
    \scriptsize
    \setlength{\tabcolsep}{4pt}
    \renewcommand{\arraystretch}{1.2}
        \centering
        \begin{tabular}{lcc}
            \multirow{2}{*}{Block} & \multicolumn{2}{c}{\oxford} \\ 
                  & (512) & (256) \\ \hline
            6  & 10.18 & 10.36 \\
            10 & 9.44  & 10.16 \\
            14 & \textbf{9.05} & \textbf{10.06} \\
            18 & 9.55  & 10.11 \\
            22 & 9.84  & 10.13 \\
        \end{tabular}
        \captionsetup{width=0.95\columnwidth}
   \caption{\footnotesize{\textbf{Label purity (\sigmalabel)} measured by computing the average standard deviation of the class labels of the top-10 most highly activating images among the top 1000 most highly activating features of the learned k-SAEs for different DiT blocks with different resolutions on \oxford. Lower is better.}}\label{tab:res}
\end{table}

%% file: tables_latex/chatgpt_prompt.tex
\begin{table}[h]
    \centering
    \begin{tcolorbox}[
        colframe=black,       
        sharp corners=southwest, 
        width=\columnwidth,   
    ]
    \begin{lstlisting}[basicstyle=\ttfamily\footnotesize, breaklines=true]
Prompt: Each set of images captures different types of patterns:
1. Class-specific information (e.g., fine-grained details, animals of the same breed).
2. Moderately granular features (e.g., similar-looking animals irrespective of their position).
3. Very coarse information (e.g., foreground objects similarly placed relative to the background).
4. Could not detect patterns (e.g., noisy or no specific patterns).
Select only one number (1, 2, 3, or 4) that best describes the shared pattern
**Respond with just the number and nothing else.**
    \end{lstlisting}
    \end{tcolorbox}
    \captionof{table}{Input prompt for GPT-4o based evaluation.}
    \label{tab:prompt}
\end{table}